\documentclass[]{interact}

\usepackage{amssymb}
\usepackage{hyperref}
\usepackage{caption, subcaption}
\usepackage{mathtools}
\usepackage{array, booktabs}
\usepackage{tabularx}
\usepackage{threeparttable}
\usepackage{algorithm}
\usepackage{algorithmicx}
\usepackage{algpseudocode}
\usepackage{makecell}
\newcolumntype{Y}{>{\centering\arraybackslash}X}
\usepackage{graphicx}
\usepackage{setspace}

\usepackage{natbib}
\bibpunct[, ]{(}{)}{;}{a}{}{,}
\renewcommand\bibfont{\fontsize{10}{12}\selectfont}

\theoremstyle{plain}

\theoremstyle{definition}

\theoremstyle{remark}

\begin{document}

\articletype{Original article}

\title{Eco-driving for Electric Connected Vehicles at Signalized Intersections: A Parameterized Reinforcement Learning approach}

\author{
\name{Xia Jiang\textsuperscript{a} Jian Zhang\textsuperscript{a}\thanks{CONTACT Jian Zhang. Email: jianzhang@seu.edu.cn} and Dan Li\textsuperscript{b}}
\affil{\textsuperscript{a}School of Transportation, Southeast University, Nanjing, Jiangsu, China; \textsuperscript{b}School of Engineering, Tibet University, Tibet, China}
}

\maketitle

\begin{abstract}
This paper proposes an eco-driving framework for electric connected vehicles (CVs) based on reinforcement learning (RL) to improve vehicle energy efficiency at signalized intersections. The vehicle agent is specified by integrating the model-based car-following policy, lane-changing policy, and the RL policy, to ensure safe operation of a CV. Subsequently, a Markov Decision Process (MDP) is formulated, which enables the vehicle to perform longitudinal control and lateral decisions, jointly optimizing the car-following and lane-changing behaviors of the CVs in the vicinity of intersections. Then, the hybrid action space is parameterized as a hierarchical structure and thereby trains the agents with two-dimensional motion patterns in a dynamic traffic environment. Finally, our proposed methods are evaluated in SUMO software from both a single-vehicle-based perspective and a flow-based perspective. The results show that our strategy can significantly reduce energy consumption by learning proper action schemes without any interruption of other human-driven vehicles (HDVs).
\end{abstract}


\section{Introduction}
\label{sec:introduction}

Transportation system has been recognised as one of the major sources of energy consumption and air pollution. The authorities have tried their utmost to build a sustainable and efficient transportation system by applying advanced technologies. One promising way to facilitate the progress is to increase the market penetrating rate (MPR) of electric vehicles with effective energy management strategies. On one hand, the electrification of vehicles promote the use of cleaner energy, which is of benefit to achieve low-emission outcomes \citep{LI2019714}. On the other hand, making full use of existing energy can help reduce energy consumption during the operation period of the vehicles. As an energy efficient purpose is inseparable from advanced technologies, the application of vehicle-to-everything (V2X) communication to traffic control and management have triggered a possible revolution towards an ecological and efficient transportation system. A plenty of research-based evidences can be found to support the view that such connected technologies are promising to enhance traffic performance \citep{JIA2016172, TALEBPOUR2016143, ARD2021103168, SHI2021103421, Xia2022}. Based on the belief that the communication modules will be deployed simultaneously on both vehicles and traffic infrastructures , the vehicle-to-infrastructure (V2I) technique can be widely implemented in the near future, which enables the connected vehicles (CVs) to obtain the information transmitted from traffic infrastructures such as traffic lights and ramp meterings, with the intent to help CVs make better decisions to improve energy efficiency.

One of a successful applications of V2I communication in urban traffic is the deployment of signal phase and timing (SPaT) message, furnishing CVs with signal status information in a consistent manner \citep{JTEPBS0000318}. Since traffic lights in urban signalized intersections can interrupt the operation of vehicles, CVs may avoid the red phases and achieve energy-efficient driving through the use of real-time SPaT data, while idling at the traffic signals is one major cause of increased energy consumption and greenhouse gas emissions\citep{KIM2020102464}. Given this, the integration of SPaT information and advanced control policies initiate eco-driving concept, which has potential to significantly reduce energy consumption in the proximity of signalized intersections \citep{LI2018335}. Generally, the control policies take vehicle location, SPaT data, and local traffic state as input and generate speed profile for the controlled vehicle (i.e., the ego vehicle), and thus enhance the performance of the vehicle in terms of energy conservation \citep{5454336, Mintsis2021, 9416184, MA2021102746}.

Eco-driving strategy at signalized intersections is usually accomplished by keeping up a smooth driving speed, refraining from aggressive acceleration, and avoiding long-time idling at the intersections, resulting in higher performance in energy economy and travel efficiency \citep{GAO2019823, 8998132}. By drawing upon the V2I communication, relevant strategies place value on the design of control algorithms in assorted traffic contexts, aiming at explicitly optimize the energy-saving speed profiles of vehicles. As such, the existing eco-driving strategies can be fundamentally classified into three broad categories: rule-based strategy, optimization-based strategy, and learning-based strategy. The mathematical formulation of each kind of strategy is different, but energy-saving performances with varying degrees are reported for all of the above methods.

Predefined rules are provided for a CV in rule-based strategies when the ego vehicle approaching the intersection. The basic ideas of rule-based methods are similar: they all divided the state into several categories according to the signal status and vehicle status, provided that the SPaT messages are available from V2I communication \citep{6083084, Xia2013, CI2019672}. As a result, the trajectory of the ego vehicle can be simply planned with classical kinematics formulas, trying to let the vehicle cross the junction within green phase. 

An advantage of the rule-based methods is that they are easy to be implemented. However, the approaches cannot adapt to dynamic traffic, because it is difficult to take the surrounding vehicles into account. For example, if the vehicle ahead of the ego CV slow down unexpectedly or the vehicle in adjacent lane insert to the placement in front of the ego CV, the rule-based methods may lose their efficacy, whereas some of them only consider the signal status and the vehicle status. Moreover, the rule-based nature is prone to lead local optimum as the speed change rate is usually fixed. In this case, the results can leave much to be desired.

In order to achieve better performance, the optimization-based approaches are extensively studied. One representative way in optimization-based strategies is to formulate an optimal control model and generate corresponding speed profiles, which is usually attained by pontryagin’s minimum principle \citep{JIANG2017290} or dynamic programming \citep{5663859, FENG2018364}. Since it is difficult to deal with the nonlinearness, time varying and environment uncertainty in such vehicle control problems, some studies adopt model predictive control (MPC) to obtain near-optimal control laws \citep{5454336,7313299,ZHAO2018802,WANG2020115233}. Another way is establishing a stage-wise approximation mathematical programming model and solving the model through heuristic algorithms, thereby providing an optimal speed control strategy. \citet{Chen2014} developed an optimization model to determine the optimal longitudinal trajectory, which is resolved by a genetic algorithm (GA). \citet{LI2018335} applied a hybrid algorithm merging GA and particle swarm optimization (PSO) to solve the multi-stage optimization model. Whereas the analytical solving process of a complex mathematical program model is time-consuming, the MPC approaches and heuristic algorithms try to come to an equitable compromise among optimality and time cost. 

There is no doubt that the optimization-based strategies have shown great potential to implement eco-driving control. Nevertheless, there still exists some major issues in this kind of strategies. Firstly, solving the mathematical models in each time step will impose a heavy burden on the on-board calculation units. The strategies may have limited applicability if the algorithm cannot be real-time. Secondly, it is hard to consider multiple vehicle behaviors (e.g., car following, lane changing, overtaking, etc) by adding a series of cost function items and constraints to the model, while a complex model can be too burdensome to be numerically solved. Thus, more adaptive strategy is supposed to be implemented to meets the requirements of real-time and dynamic.

Recently, with the rapid development of artificial intelligence, the reinforcement learning (RL) algorithms have attracted the gaze of researchers in both optimal control community and transportation community. Through conducting the learning procedure, an agent can select an optimal action for each observation to maximize its cumulative expected reward (i.e., usually is the optimization goal), and this process can be usually achieved in a real-time fashion \citep{CHOW2021103264}. By leveraging deep neural networks, the integrating outcome, which is known as deep reinforcement learning (DRL), has the ability to approximate the optimal policy in most of the control tasks, even with high-dimension state space . Consequently, the potent DRL-based methods are introduced to control the vehicle in the vicinity of signalized intersections.

\citet{Shi2018} used a regular Q-learning algorithm to implement eco-driving for a CV in free flow condition. They took the total CO$_2$ emission as the reward signal, aiming at optimize the driving behavior of the ego vehicle by yielding a discrete acceleration rate in each time step. As one of the value-based reinforcement learning algorithms, the Q-learning approach cannot control the vehicle in continuous acceleration space, and thereby causes local optimum and uneven trajectory in most cases. The framework developed by \citet{Mousa2020} provided insight into the DRL-based eco-driving system, which introduced deep Q network (DQN) to improve the fuel performance of the controlled CV. However, major disadvantage similar to the work of \citet{Shi2018} was encountered in their study, namely, the losing efficacy in continuous action space. As the policy-based algorithms are capable of learning control policies with continuous action sets, researchers tend to apply typical policy-based DRL algorithms to surmount the barrier of continuous space. \citet{8848852} proposed a car following model based on the deep deterministic policy gradient (DDPG) algorithm, which is proved can improve travel efficiency, fuel consumption and safety at an isolated signalized intersection. Similar study was also formulated for EV when traffic oscillations were concerned \citep{QU2020114030}. \citet{GUO2021102980} combined DDPG and DQN to formulate a hybrid reinforcement learning framework, and the developed approach can not only control the longitudinal motion, but also optimize the lateral decision of the ego vehicle. \citet{WEGENER2021102967} studied the application of twin-delayed deep deterministic policy gradient (TD3) algorithm, which conformed to the similar basic idea of DDPG, but introduced a series of tricks to tackle the Q function overestimate problem of DDPG. Their simulation study proclaimed that the DRL algorithm can adapt to the eco-driving scenarios with the presence of other surrounding vehicles.

\begin{table}\caption{The summary of existing RL-based eco-driving frameworks}
\label{tab:table1}
\footnotesize
\begin{threeparttable}
\begin{tabularx}{\textwidth}{cXXccX}
\toprule
Study & Vehicle type & RL algorithm & car follow & lane change & action space \\
\midrule
\citet{Shi2018} & GV & Q-learning & $\surd$ & - & Discrete \\
\citet{Mousa2020} & GV & DQN & $\surd$ & - & Discrete \\
\citet{8848852} & GV & DDPG & $\surd$ & - & Continuous \\
\citet{QU2020114030} & EV & DDPG & $\surd$ & - & Continuous \\
\citet{WEGENER2021102967} & EV & TD3 & $\surd$ & - & Continuous \\
\citet{GUO2021102980} & GV & DDPG+DQN & $\surd$ & $\surd$ & Hybrid \\
\citet{bai2022hybrid} & EV & Dueling DQN & $\surd$ & $\surd$ & Discrete \\
\bottomrule
\end{tabularx}
 \begin{tablenotes}
        \footnotesize
        \item[*] GV - Gasoline Vehicle; EV - Electric Vehicle
\end{tablenotes}
\end{threeparttable}
\end{table}

Focusing on the energy-efficient driving process at signalized intersections, the existing RL-based frameworks differ from each other in multiple dimensions, such as the category of the algorithm, the studied vehicle type, the driving scenario, the design of the action space, etc. \autoref{tab:table1} presents the summary of relevant studies. It is found that the lane changing behavior is rarely taken into account when developing eco-driving strategies. Lane changing during the driving process may bring about a long-term energy benefit, especially under the situations that the preceding vehicle or queue can interfere with the running of a CV. Although \citet{GUO2021102980} considered the combination of car following and lane changing as their hybrid RL-based approach simply formulated two Markov Decision Processes (MDP) for longitudinal motion and lateral decision of the ego vehicle. Therefore, they only control the two dimensional motions separately by DQN and DDPG. The inherent mechanism of the hybrid MDP is not clear, and it may lead to local optimum because the movements in two dimensions are not optimize jointly. It should be noted that the work of \citet{bai2022hybrid} also considers lane changing, but the nature of DQN determines that it cannot guarantee a smooth trajectory and corresponding global optimum, whereas the running of a vehicle should be continuous process. In this case, in order to handle the application of discrete-continuous hybrid action space, we introduce the parametrized action space and compose an unified MDP for eco-driving on the signalized urban road. Differing from traditional RL methods, the proposed parametrized reinforcement learning (PRL) controller is capable of optimizing the longitudinal motion and the lateral motion jointly and stipulate a better performance, further exploiting the energy-saving potential of electric and connected vehicles through DRL-based speed control.

In addition, since the inappropriate driving behaviors usually undermine the holistic performance of traffic, the operation of surrounding normal human-driven vehicles (HDVs) may be interrupted by the CVs controlled by eco-driving controllers \citep{ZHAO2018802}. We call this kind of strategy "selfish" eco-driving, which may caused by frequent lane changing or unexpected deceleration. Compared with most of the other learning-based strategies, which usually pay close attention to single-vehicle performance, in this paper, we will restrict CVs from frequent lane changing and unexpected change of acceleration when designing the reward function, and test whether the proposed strategy has detrimental influence on other HDVs.

In summary, the particular contributions and the novelty of this paper lie within the following points:
\begin{itemize}
    \item An MDP with discrete-continuous hybrid action space is established for CVs in signalized urban scenario, further tapping the energy-saving potential of CVs by searching in the longitudinal acceleration/deceleration space and the lateral lane changing space.
    \item Instead of simply blending algorithms or discretize the hybrid action space, the proposed PRL algorithm can be naturally applied to MDP with hybrid action space, which provides a paradigm for such microscopic CV control problems.
    \item The proposed strategy can be harmonious by designing a restricted reward function so that to eliminate the potential adverse impact of the selfish behaviors of the ego vehicles.
    \item The developed methodology is tested in both microscopic (single vehicle-based) perspective and macroscopic (mixed traffic flow-based) perspective, illuminating the energy-saving capability of the learning-based approach.
\end{itemize}

The organization of the remainder paper is as follows: The preliminary of RL is presented in Section~\ref{sec:background}. Section~\ref{sec:system architecture} introduces the system components regarding the eco-driving process of CVs based on RL controller. Section~\ref{sec:PRL for eco-driving} describes the establishment of the MDP with the implementation of the PRL algorithm. Numerical results are presented and discussed in Section~\ref{sec:result} by carrying out simulations in microscopic simulation platform SUMO \citep{SUMO2018}. Finally, some concluding remarks are provided in Section~\ref{sec:conclusions}.

\section{Background of Reinforcement Learning}
\label{sec:background}
The environment (i.e., traffic environment) of the RL agent (i.e., the ego vehicle) is modeled by an MDP $M=\{\mathcal{S},\mathcal{A},\mathcal{P},r,\gamma\}$ in reinforcement learning framework, where $\mathcal{S}$ denotes the state space, $\mathcal{A}$ denotes the action space,  $\mathcal{P}$ is the Markov transition probability distribution, $r$ is the reward function, and $\gamma \in [0,1]$ is the discount factor that measures the relative value of future rewards and current reward. 

In the general setup of RL, an agent interacts with the environment sequentially as follows. At time step $t$, the agent observes a state $s_t \in \mathcal{S}$ and selects an action $a_t \in \mathcal{A}$, and then the agent receives an immediate reward value calculated by $r(s_t, a_t)$. Subsequently, the next state becomes $s_{t+1}~P(s_{t+1}|s_t,a_t)$. Let $R_t=\sum_{j \geq  t}\gamma^{j-t} r(s_j,a_j)$ be the cumulative discounted reward starting from time step $t$, for any policy $\pi$, we define the state-action value function as $Q^{\pi}(s,a)=\mathbb{E}[R_t|s_t=s,a_t=a,\pi]$, where $a_{t+k} \sim \pi(\cdot|s_{t+k})$ for all $a_{t+k}$ and $s_{t+k}$ for $k \in [t+1,\infty)$. Meanwhile, we define the state value function as $v^{\pi}(s)=\mathbb{E}[R_t|s_t=s,\pi]$. According to the Bellman equation, we have $v^{\pi}(s)=\sum_{a \in \mathbb{A}}\pi(a|s)Q^{\pi}(s,a)$, which illuminates the relationship between the two function. The goal of the agent is to learn the optimal policy $\pi^*$ that can maximize its expected cumulative discounted reward from the initial state, which can be represented as $J(\pi)=\mathbb{E}(R_0|\pi^*)$, while the learning 
procedure is usually accomplished by estimating the optimal state-action value function $Q^{\pi^{*}}$.

Value-based methods and policy-based methods are two categories of RL algorithms, the former estimates $Q^{\pi^{*}}$ and generates actions in a greedy way, while the latter directly learns the optimal policy $\pi^{*}$ by optimizing $J(\pi)$. As one typical representative of value-based methods, the Q-learning algorithm \citep{watkins1992q} is deduced by Bellman equation:
\begin{equation}\label{eq:eq1}
    Q(s,a) = \underset{r_t,s_{t+1}}{\mathbb{E}} [r_t+\gamma \max_{a^{'} \in \mathcal{A}} Q(s_{t+1}, a^{'})|s_t=s,a_t=a]
\end{equation}
which takes $Q^{\pi^{*}}$ as the unique solution. The learning process iteratively updates the function $Q(s,a)$ through Monte-Carlo method over a series of sample transitions. Conventionally, $Q(s,a)$ is stored in tabular data structure for finite state space configuration. Nevertheless, the problem of large state space arises along with the increase of the task complexity, When the state space is too large to be stored in computer memory or $Q$-value cannot be retrieved in real time, the 
function approximation approach is introduced, especially promoted by the development of deep neural networks. DQN utilizes a deep neural network $Q(s,a;\theta) \approx Q(s,a)$ with parameter set $\theta$ to approximate $Q^{\pi^{*}}$ \citep{mnih2015human}. The learning of the agent is fundamentally based on the gradient descent theory, which takes the following loss function:
\begin{equation}\label{eq:eql}
    L_t(\theta)=\{Q(s_t,a_t;\theta)-[r_t+\gamma \max_{a^{'} \in \mathcal{A}} Q(s_{t+1}, a^{'};\theta_t)]\}^2
\end{equation}

Differing from the value-based approaches, the policy-based algorithms with deep neural network directly model the parametrized policy $\pi_{\theta}$. The parameter set $\theta$ is updated iteratively via gradient descent to maximize $J(\pi_{\theta})$. A typical application is REINFORCE algorithm, which takes the corresponding loss function:
\begin{equation}
    \Delta_\theta J(\pi_\theta) = \underset{s_t,a_{t}}{\mathbb{E}} [\Delta_\theta log \pi_\theta(a_t|s_t)Q^{\pi_\theta}(s_t,a_t)]
\end{equation}

\section{Methodology}
\label{sec:system architecture}

\subsection{Scenario description}
In this paper, we consider a fully electric vehicle environment to better measure the holistic energy performance of the mixed road traffic. which  is composed of HDVs and CVs in urban scenario, while only CVs can receive the SPaT data and be controlled by eco-driving controller. By incorporating such a mixed traffic condition, the developed strategy is closer to the urban traffic in the near future so that it can be more practical. 

\begin{figure}[th]
    \centering
    \includegraphics[width=0.95\textwidth]{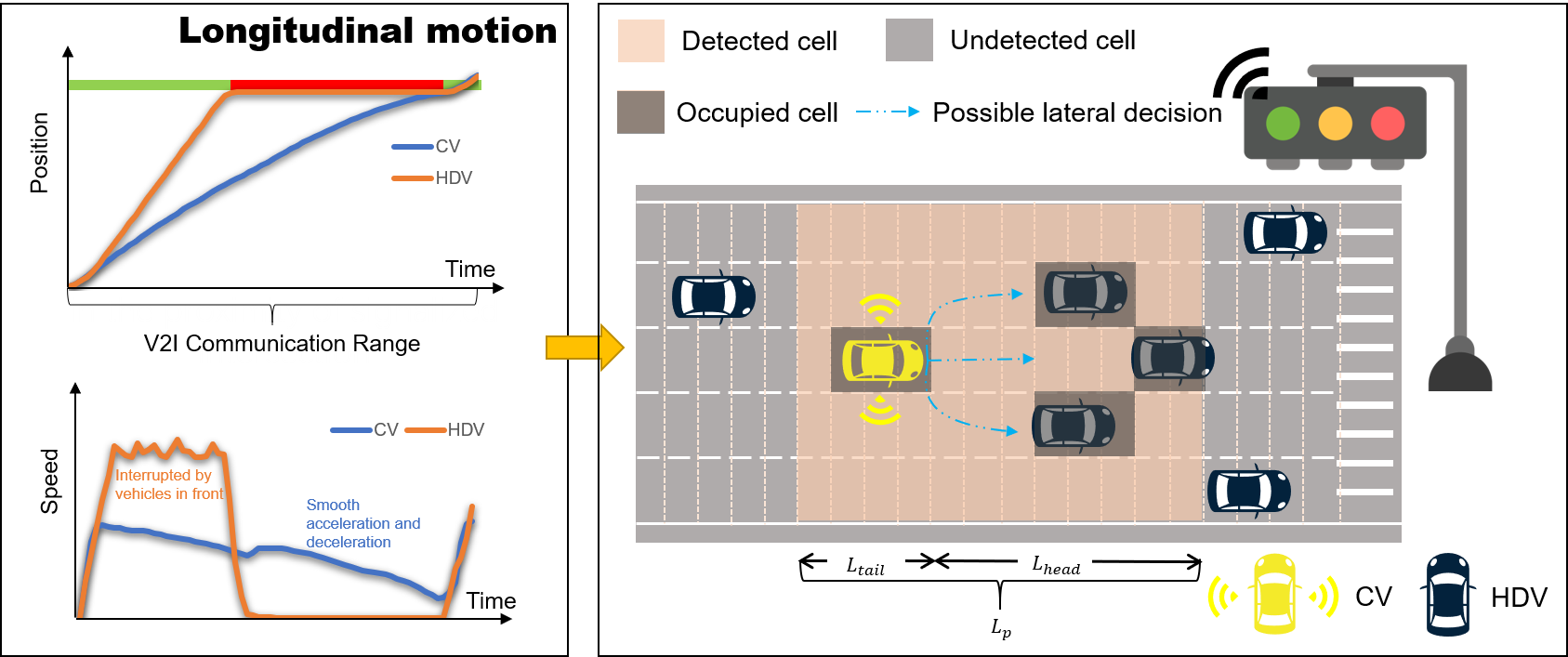}
    \caption{The driving scenario for controlled vehicle in the proximity of a signalized intersection.}
    \label{fig:framework}
\end{figure}

\autoref{fig:framework} shows the general eco-driving process of the ego vehicle. The left half of the figure depicts a typical longitudinal trajectory and speed profile of a CV with eco-driving system. Compared with HDVs, the CV with eco-driving controller can generate more smooth speed profile and navigate the vehicle cross the junction in green phase, in order to minimize its electricity consumption. Considering a more specific scenario, which is shown in the right half of \autoref{fig:framework}, the CV on the road can constantly detect the surrounding traffic state within its perceived range. Inspired by \citep{9325948}, we define the perceived range of the CV as a two-dimensional occupancy grid. The grid cells reflect the occupancy situation of corresponding road segments around the CV, so we call the grid local traffic state. Whereas the CVs are equipped with multiple sensor devices, such as camera, Lidar, and radar, this process can be certainly achieved. In addition to local traffic state, the CV could also acquire the SPaT data based on V2I communication, provided that it enters the V2I communication range. The local traffic state and SPaT-related features are expected to serve as the state of the CV agent, which is one of the main components in the MDP model.

According to the position of the vehicle, the operation of the CV is categorized into two cases: (1) When the position of the CV is not within the V2I communication range of the first downstream traffic signal, it operates by leveraging local traffic state and default value of the SPaT information. This is also true of the circumstance when the ego vehicle is inside the intersection for a multi-intersection scenario. (2) During the period that the V2I communication is available for the ego CV, it continuously acquire traffic signal state, together with  local traffic state, to adjust its longitudinal acceleration and make proper lateral decision under control of the eco-driving system. For longitudinal dimension, the ego vehicle is supposed to cruise with smooth speed profile and crossing the intersection in green phase; for lateral dimension, the ego vehicle can stay at current lane or change to adjacent lane to mitigate the interruption caused by other HDVs or vehicle queues.

\subsection{Agent framework}
The CVs are deemed as agents, which are supposed to adapt to the dynamic traffic environment and make proper decisions. An agent is a combination of a PRL policy and several model-based policies. The motivation of the hybrid policy is: (1) The DRL algorithms always suffer from credibility due to its unexplainability, and this may cause issues related to driving safety, while model-based policies explicitly constraint the behavior of vehicles to ensure a collision-free driving. (2) The model-based driving policies cannot react to traffic environment adaptively with pre-defined rules, while DRL policy with "trial-and-error" mechanism can explore the action space more effectively. This hybrid form endows the agent with the ability to cope with complex driving scenarios on the premise of safety. 

\begin{figure}[!h]
    \centering
    \includegraphics[width=0.95\textwidth]{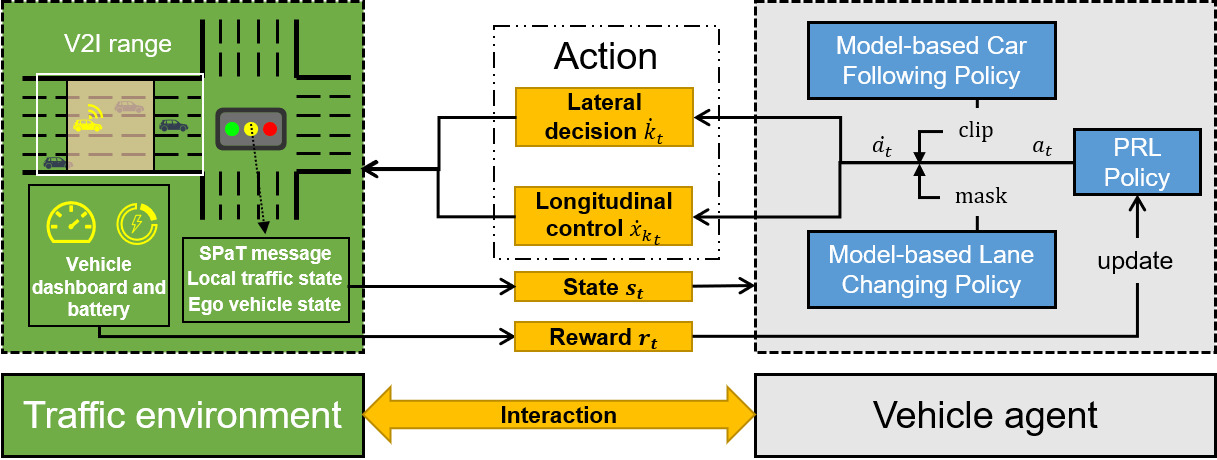}
    \caption{The workflow of the agent when V2I communication is available.}
    \label{fig:agent}
\end{figure}

When the agent enters the V2I communication range, its workflow can be seen in \autoref{fig:agent}. In each step, the agent can observe a state $s_t$ and perform action $a_t$ which is output by the PRL policy, and subsequently receive a reward value $r_t$. We abstract the lateral decision and the longitudinal control of the CV as discrete process and continuous process, respectively. The action $a_t$ thereby can be defined as a tuple $(k_t, x_{k_t})$, where $k_t$ is an integer value that represents the possible lateral decision that is shown in \autoref{fig:framework}, and $x_{k_t}$ is an float value that represents longitudinal acceleration.

During the learning process of the agent, rear-end or lateral collision may be caused by unreasonable actions. Concerning that the original action generated by the PRL policy $a_t=(k_t, x_{k_t})$ may be unsafe, we propose the clip operation and mask operation to preprocess the original action $a_t$, and then decompose the processed action $\overset{.}{a_t}$ to perform reliable motion control in both longitudinal and lateral dimension. This facilitates more efficient agent training as apriori knowledge is added to prevent the vehicle from unexpected collision.

In order to provide full explanation of how the framework works, we need to specify the formulation of the MDP for the agent:

Firstly, the state space $S$ of the agent comprises two parts, namely, the local traffic state and the logic state, which are specified as:

\begin{itemize}
    \item Local traffic state $M_t^i$ for agent $i$ at time $t$ is defined as a two-dimensional matrix that reflects the occupancy situation of the perceived range of the agent, which is illustrated in \autoref{fig:framework}. We assume the perceived range of the CV is $L_p$, and it can be divided into forward range $L_{head}$ and backward range $L_{tail}$. The forward range $L_{head}$ can help the CV maintain a reasonable speed, when the vehicle agent can obtain the relative distance between itself and the front vehicle. The backward range $L_{tail}$ can prompt the CV to find a reasonable lane-changing interval that can be safe and harmonious. The shape of the local traffic state matrix is $(\frac{L_p}{L_c}, 5)$ for a 5-lane road with the cell length $L_c$. The elements in the matrix are filled with "0" if the associated detected cells are occupied by vehicles, otherwise they are filled with "1". 
    \item Logic state $V_t^i$ for agent $i$ at time $t$ is defined as a vector that includes the physical state of the ego vehicle and the state of the traffic signal. The logic state of the agent is denoted as $V_t^i=[E_t^i, d_t^i, v_t^i, f_t^i, g_t^i]^T$, where $E_t^i$ is the one-hot encoding of the lane index that the ego vehicle $i$ located; $d_t^i$ describes the distance between agent $i$ and the downstream traffic signal at time $t$; $v_t^i$ is the velocity of the agent at time $t$; $f_t^i$ is a boolean variable that indicates whether the next traffic signal is in red phase; $g_t^i$ denotes the time duration between time $t$ to the time that the next traffic signal turn to green phase for the agent. When V2I communication is unavailable, the value of $f_t^i$ and $g_t^i$ are all set to default value -1.
\end{itemize}

As a result, the state in time $t$ for agent $i$ is defined as $s_t^i=(M_t^i, V_t^i)$, in which the former element is expected to be processed by a convolutional neural network block to extract latent features, and the latter element will concatenate to the extracted local traffic state features and be input to a multi-layer perceptron block to train the PRL policy.

Secondly, the action space in this paper is considered as a hybrid form that support both longitudinal control and lateral decision. As mentioned above, the basic representation of the action at time $t$ is $a_t = (k_t, x_{k_t})$, where $k_t \in \{-1, 0, 1\}$ and $x_{k_t} \in [a_{min}, a_{max}]$. The value set $\{-1, 0, 1\}$ of $k_t$ represent that the ego vehicle can change to the left lane, stay at the current lane, and change to the right lane. We assume that the lane changing process can be finished in a fixed duration $T_c$. The space of $x_{k_t}$ limits the acceleration of the CV to the range of its maximal deceleration $a_{min}$ and maximal acceleration $a_{max}$. Instead of combining discrete speed variation rate and lane changing decision to formulate an overall discrete action space, we utilize the hybrid discrete-continuous action space to effectively explore the solution space and obtain better performance. Since imposing the original $a_t$ on the ego vehicle may make it be exposed to traffic accident, we introduce conventional car-following model and lane changing model to modify the dangerous driving behavior. In this paper, we utilize the Krauss car-following model \citep{etde_627062} and the "LC2013" lane-changing model \citep{dlr102254}, which are embedded in SUMO simulation software, to perform clip and mask operations and obtain the modified action $\overset{.}{a_t}=(\overset{.}{k_t}, \overset{.}{x_{k_t}})$:

\begin{itemize}
\item Clip operation aims at clipping the potential risking acceleration to a safe value to avert rear-end collisions or beyond the road speed limit $V_{max}$. We assume that the acceleration output by the utilized car-following model is $x_t^{CF}$ at time $t$ according to current driving status, and then the corrected acceleration is transformed to $\overset{.}{x_{k_t}}=\min(x_{CF}^t,x_{k_t})$. Hence, the velocity of agent $i$ at time $t$ is clipped as $v_t^i=\max(\min(V_{max},v_{t-1}^i+\overset{.}{x_{k_t}}),0)$, which confines the velocity of the CV to a reasonable range.

\item Mask operation is used to judge whether the lateral decision produced by PRL policy is safe, in order to prevent lateral collisions or deadlocks. For each non-zero discrete action $k_t$, the model-based lane changing policy output an bool value $\xi$ to resolve that if the agent could change lane. The corrected lateral action is denoted as $\overset{.}{k_t}=k_t \times \xi$.
\end{itemize}

Thirdly, we discuss different settings of the reward function $r_t$, while different reward signal may trigger various learning directions of the agent. The general goal of the eco-driving strategy is to make a trade-off between energy consumption and travel time, as the energy-efficient driving mode may cause the increase of the delay \citep{WANG2020102188}. An intuitive way to set the reward function is simply returning a linear combination of total electricity consumption $E_f$ and travel time $T_f$:

\begin{equation}
    r_t^1 = \begin{cases}
    -(E_f + \varphi T_f) , \mbox{if } t=t_f \\
    0, \mbox{otherwise}
    \end{cases}
\end{equation}
where, $\varphi$ is a weighting parameter to measure the relative importance of energy indicator and mobility indicator; $t_f$ is the terminal time step (i.e., the time that the ego vehicle leaves the junction).

However, the travel time can only be calculated at the end of the travel, and it is difficult for the vehicle agent to learn effectively from such a delay reward signal. Given this, \citet{GUO2021102980} used a step-wise travel distance $l_t^i$ as the proxy of the travel time. Incorporating with the step-wise energy consumption $e_t^i$, the instantaneous reward function is defined as:

\begin{equation}
    r_t^2 = l_t^i - \varphi e_t^i
\end{equation}

Despite signs of an improvement in the energy-saving performance, we find that it is hard to generate a smooth trajectory for the agent by deploying $r_t^2$. Meanwhile, this design can be defective because the agent tends to maximize its future cumulative travel distance, while the remaining distance from the location of the CV to the intersection is fixed so that the overall travel time cannot be well reflected. Thus, we devise a novel reward function composed of step-wise reward and terminal reward to facilitate the progress of the eco-driving task in a hybrid action space. It is based on the following considerations:

\begin{itemize}
    \item The velocity is one of the most important factors when we place value on the mobility of the CVs, when continuously low speed will lead to an increase of travel delay. Thus, we set a penalty item with a constant value $r_p^s$ when the velocity of the agent is lower than a threshold value $v_{\zeta}$.
    \item Frequent changes of acceleration can cause traffic oscillation by fluctuating the speed profile of vehicles, and impair the driving comfort at the same time \citep{ZHAO2018802}. The differential form of the acceleration is usually called jerk, which manifests the variation rate of the acceleration. Given this, we set another penalty item $r_p^j$ when the value of jerk is higher than a threshold value $j_{\zeta}$.
    \item Since frequent lane changing will interfere with the normal operation of other CVs or HDVs, we set a harmony coefficient $r_p^c$ for to penal each lane changing behavior of the agent so that it can change lane in necessary cases.
\end{itemize}

The three penalty items instruct the agent maintain a smooth speed profile with as little stopping and lane changing as possible. Finally, our reward function is defined in \autoref{eq:rwd}. It should be noted that the calculation of $E_f$ and $T_f$ is only triggered when the episode reaches the terminal state.

\begin{equation}\label{eq:rwd}
    r_t = r_p^s + r_p^j + r_p^c - (E_f + \varphi T_f)
\end{equation}

\section{PRL for eco-driving}
\label{sec:PRL for eco-driving}

\subsection{Reinforcement Learning in Hybrid Action Space}
In addition to normal discrete control and continuous control, which are widely studied by previous works, some control tasks intrinsically have hybrid action spaces. Controlling of CV is one representative in this field, as the vehicle can move longitudinally and laterally on the road. While the majority of the existing studies are confined to car following motion \citep{Shi2018,Mousa2020,8848852,WEGENER2021102967}, this paper develops a parameterized action space to naturally describe the control problem with hybrid actions and thereby implement joint optimization of car-following and lane-changing movement.

\begin{figure}[ht]
    \centering
    \includegraphics[width=0.95\textwidth]{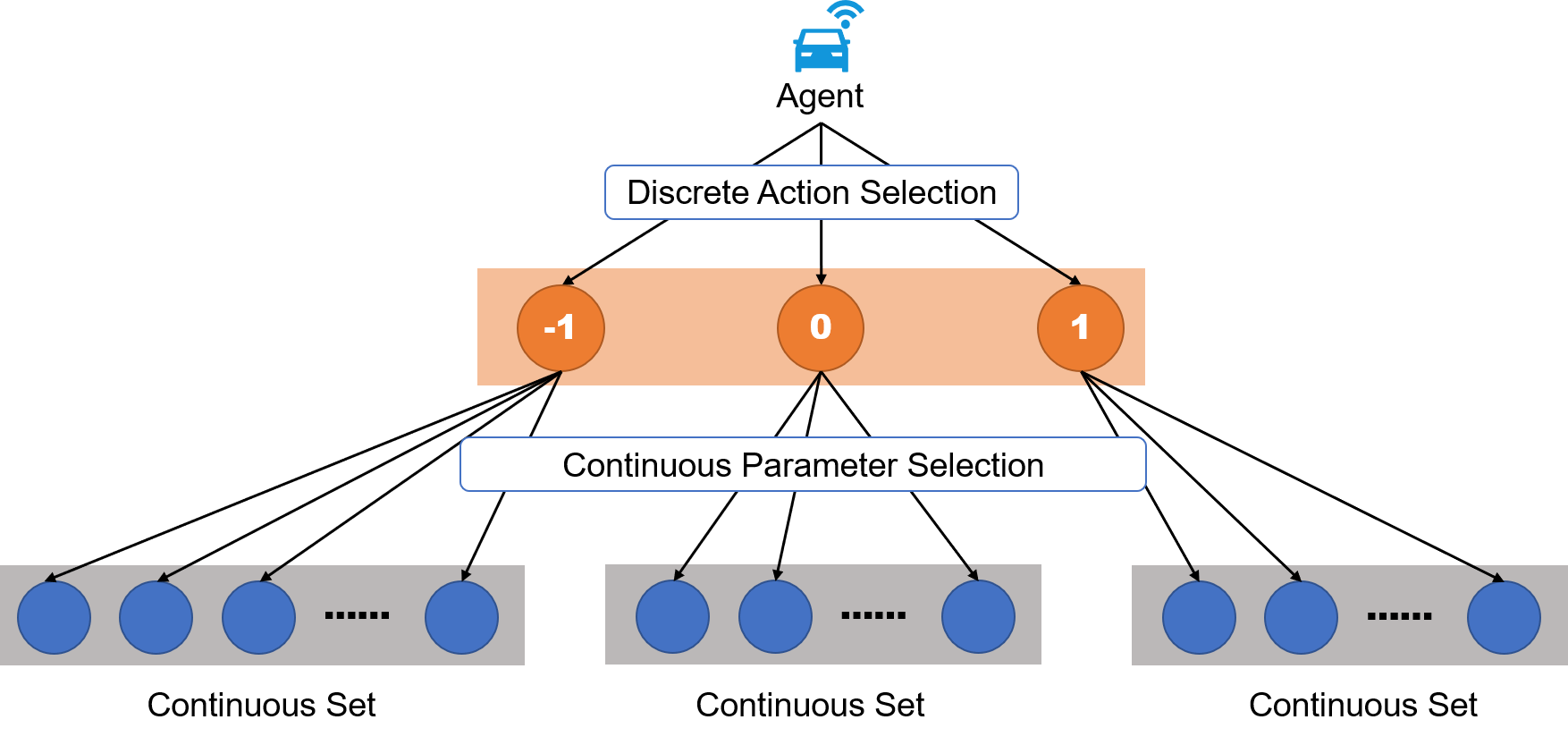}
    \caption{The illustration of a parameterized action space.}
    \label{fig:fig1}
\end{figure}

\autoref{fig:fig1} presents an example of parameterized action space. We assume that the lane changing process of the vehicle agent is controlled in a discrete space (i.e., changing to the left lane, staying at current lane, changing to the right lane), which is marked with rectangles in orange. Each discrete action has a continuous parameter space (i.e., the longitudinal acceleration/deceleration) marked with rectangles in grey. The hierarchical structure formulates the parameterized action space in which the agent select actions.

Considering an MDP with parameterized action space, the agent first selects a discrete action $k_t \in \mathcal{K}$ at time $t$, and then selects an associated continuous parameter $x_{k_t} \in \mathcal{X}_k$. In this case, the Bellman equation can be transformed to:
\begin{equation}\label{eq:eq4}
    Q(s_t,k_t,x_{k_t})=\underset{r_t,s_{t+1}}{\mathbb{E}} [r_t+\gamma \max_{k \in \mathcal{K}} \sup_{x_k \in \mathcal{X}_k} Q(s_{t+1}, k,x_k)|s_t=s,a_t=(k_t,x_{k_t})]
\end{equation}

Note that searching for an optimal action parameter $x_k$ over continuous space in each time step is computationally intractable. Therefore, by setting up a function $x_k^Q: \mathcal{S} \rightarrow \mathcal{X}_k$ to represent $\arg \sup_{x_k \in \mathcal{X}_k} Q(s,k,x_k)$, the Bellman equation can be rewritten as \autoref{eq:eq5} \citep{xiong2018parametrized}.

\begin{equation}\label{eq:eq5}
    Q(s_t,k_t,x_{k_t})=\underset{r_t,s_{t+1}}{\mathbb{E}} [r_t+\gamma \max_{k \in \mathcal{K}} Q(s_{t+1}, k, x_k^Q(s_{t+1}))|s_t=s]
\end{equation}

Given this, a deep neural network with parameter set $\theta$ can be used to approximate Q function: $Q(s,k,x_k;\theta) \rightarrow Q(s,k,x_k)$. For such a $Q(s,k,x_k;\theta)$, another neural network with parameter set $\omega$ can serve as the a deterministic policy network to approximate $x^Q_k(s)$, that is, $x_k(:|\omega): \mathcal{S} \rightarrow \mathcal{X}_k$. When $\theta$ is determined, the purpose of the algorithm is to find $\omega$ such that

\begin{equation}\label{eq:eq6}
    Q(s,k,x_k(s;\omega);\theta) \approx \sup_{x_k \in \mathcal{X}_k} Q(s,k,x_k;\theta)  \mbox{\quad for each \quad} k \in \mathcal{K}
\end{equation}

Subsequently, we can estimate $\theta$ by minimizing the mean-squared Bellman error via typical gradient descent, and this resembles the learning process of DQN. The formation of the loss function for $\theta$ is similar to \autoref{eq:eql}:

\begin{equation}\label{eq:eq7}
    L_t^Q(\theta) = \frac{1}{2}[Q(s_t,k_t,x_{k_t};\theta) - y_t]^2
\end{equation}
where, $y_t$ denotes the cumulative future reward $r_t+\gamma\max_{a^{'} \in \mathcal{A}}Q(s_{t+1},a^{'};\theta_t)$ in time $t$. When $\theta$ is fixed, we aim to regulate $\omega$ to maximize $Q(s,k,x_k(s;\omega))$, so the loss function for $\omega$ is as following

\begin{equation}\label{eq:eq8}
    L_t^{\omega}(\omega) = -\sum^{\mathcal{K}}_{k=1}Q(s,k,x_k(s;\omega);\theta_t)
\end{equation}

We thereby deduce the reinforcement learning theory in a hybrid action space. To sum up, the framework utilizes a parameterized action space to integrate the hybrid action sets in a hierarchical fashion. Then we formulate two neural networks to learn the optimal discrete action and its associated continuous action, respectively.

\subsection{PRL algorithm}
The DRL algorithm with parameterized action space should not only accommodates the eco-driving environment, but also considers the implementation details, such as the exploration process, the design of replay buffer, the setting of target networks, to make the training process effective and stable. 

\begin{figure}[ht]
    \centering
    \includegraphics[width=0.95\textwidth]{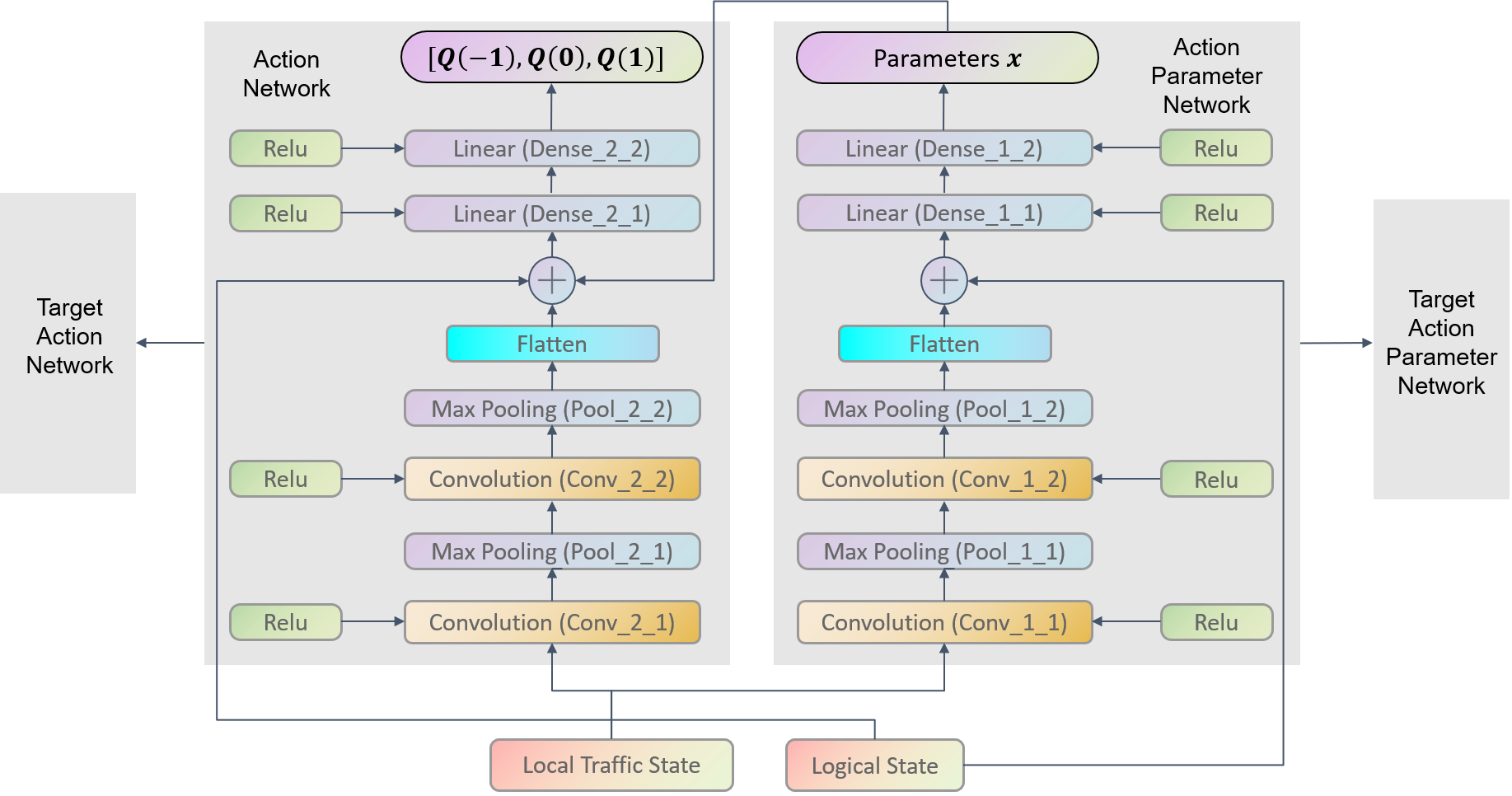}
    \caption{The architecture of the policy network.}
    \label{fig:fig2}
\end{figure}

Firstly, the architecture of the policy network for the agent should be specified by several neural networks, which is shown in  \autoref{fig:fig2}, in order to formulate the backbone of the agent. It takes an action parameter network with weights $\omega$ to output the parameter vector $x$, and uses an action network with weights $\theta$ to approximate the optimal parameterized Q function. For each neural network, the local traffic state is taken as the input of a convolutional neural network to extract latent features, and then the flatten features is concatenated with logical state. For the action network, the parameter vector $x$, which is produced by action parameter network, is fitted together in the concatenation operation. After a series of fully-connected layers, the corresponding output is obtained.

In addition, target network is introduced for both action parameter network and action network to curd instability during training. A target network is a copy of the original network that is held fixed to serve as a stable target for some number of steps. We update the target network in a soft way:

\begin{equation}\label{eq:tau1}
    \theta^{'} = \tau_1\theta + (1-\tau_1)\theta^{'}
\end{equation}

\begin{equation}\label{eq:tau2}
    \omega^{'} = \tau_2\omega + (1-\tau_2)\omega^{'}
\end{equation}
where, $\theta^{'}$ and $\omega^{'}$ are the parameter set of target action network and target action parameter network; $\tau_1$ and $\tau_2$ are hyperparameters to determine the averaging extents.

As both longitudinal decision and lateral decision should be optimized, it is necessary to perform exploration for the hybrid action set during training so that more effective motion patterns can be searched. We deploy two exploration strategies for the agent, namely, $\epsilon \text{-greedy}$ strategy and Ornstein–Uhlenbeck process noise. The former strategy is usually utilized in value-based DRL algorithm, similar to the DQN approach, while the latter resembles that in DDPG algorithm. Meanwhile, a linear decay method is achieved to regulate the trade-off between exploration and exploitation over the whole training process with numerous episodes. In this case, the formulation of the decaying $\epsilon \text{-greedy}$ strategy is

\begin{equation}\label{eq:k}
    k = \begin{cases}
                \arg\max_{k \in \mathcal{K}} Q(s, k, x_k(s)),  \mbox{with probability } (1-\epsilon) \\
                exploration, \mbox{with probability } \epsilon
        \end{cases}
\end{equation}
where, $exploration$ represents a randomly selected discrete action. The value of $\epsilon$ is determined by

\begin{equation}\label{eq:explore2}
    \epsilon = (\epsilon_{init}-\epsilon_{end})\max(\frac{N-n_{step}}{N},0)+\epsilon_{end}
\end{equation}
where, $\epsilon_{init}$ and $\epsilon_{end}$ are the initial value and the terminal value of exploration factor $\epsilon$ during the overall learning episodes; $N$ denotes the total decaying episodes; $n_{step}$ denotes the index of current episode.

Ornstein–Uhlenbeck process \citep{PhysRev36823} is implemented to generate temporally correlated noise, which embodies the exploration mechanism for action parameter network. By adding noise sampled from a noise process $\mathcal{N}$, the continuous action produced by action parameter network $\mu(s|\omega)$ can be slightly disturbed, that is

\begin{equation}\label{eq:explore3}
    x_k = \mu(s|\omega) + \mathcal{N}_k
\end{equation}

The holistic process of the proposed PRL algorithm is presented in Algorithm 1. Before the main loop, the weights of the action network, action parameter network, and the target networks are initiated. For each episode of training, line 3 shows the initial operation of the simulation traffic environment. The initialization of the simulation should consider different introductory states to enhance generalization ability of the agent, and this will be introduce in \autoref{sec:result}. The updating of the weights of the neural networks follows line 17 and line 18 in Algorithm 1. By iteratively updating the parameters, the ego vehicle, which is seen as the training agent, is capable of conforming to the dynamic traffic environment and the changing of downstream traffic signals.

\begin{algorithm}[htp]\label{alg:algo1}
\caption{The PRL algorithm for eco-driving agent training}
\hspace*{0.02in}{\bf Hyperparameters:}
averaging rate $\tau_1$ and $\tau_2$, exploration factor $\epsilon$, initial exploration factor $\epsilon_{init}$, terminal exploration factor $\epsilon_{end}$, the number of decaying episodes $N$, stepsizes $\{\alpha_t,\beta_t\}_{t \geq 0}$, and minibatch size $B$ \newline
\hspace*{0.02in}{\bf Initialize}:
action network $Q(s,k,x_k|\theta)$ and action parameter network $\mu(s|\omega)$, target action network $Q^{'}$ with weights $\theta^{'} \leftarrow \theta$ and target action parameter network $\mu^{'}$ with weights $\omega^{'} \leftarrow \omega$, replay buffer $\mathcal{R}$, $n_{step} \leftarrow 0$
\begin{algorithmic}[1]
\For{each episode}
    \State $n_{step} \leftarrow n_{step} + 1$
    \State loading initial traffic environment and observe current state $s_t$
    \If{$n_{step} < N$}
        \State update exploration factor $\epsilon$ according to \autoref{eq:explore2}
    \EndIf
    \State initialize Ornstein–Uhlenbeck process $\mathcal{N}$ for continuous action exploration
    \For{each step in simulation}
        \State compute action parameters $x_{k_t}$ according to \autoref{eq:explore3}
        \State compute action $k_t$ according to \autoref{eq:k}
        \State take action $a_t=(k_t,x_{k_t})$, observe reward $r_t$ and next state $s_{t+1}$
        \State store transition $\{s_t, a_t, r_t, s_{t+1}\}$ into $\mathcal{R}$
        \State sample $\mathcal{B}$ transitions $\{s_b, a_b, r_b, s_{b+1}\}_{b\in\mathcal{B}}$ randomly from $\mathcal{R}$
        \State decompose $a_b$ into $k_b$ and $x_{k_b}$
        \State compute future cumulative reward $y_b$ for batch $\mathcal{B}$ based on $Q^{'}$ and $u^{'}$
        \State use data $\{y_b,s_b,k_b\}_{b\in\mathcal{B}}$ and $\{y_b,s_b,x_{k_b}\}_{b\in\mathcal{B}}$ to compute the stochastic gradient $\nabla_\theta L_t^Q(\theta_t)$ and $\nabla_\omega L_t^{\omega}(\omega_t)$
        \State update $\theta$ and $\omega$ by $\theta_{t+1} \leftarrow \theta_t - \alpha_t\nabla_\theta L_t^Q(\theta_t)$ and $\omega_{t+1} \leftarrow \omega_t - \beta_t\nabla_\omega L_t^{\omega}(\omega_t)$
        \State update the parameters of target neural networks according to \autoref{eq:tau1} and \autoref{eq:tau2}
    \EndFor
\EndFor
\end{algorithmic}
\end{algorithm}

The hyperparameters of the algorithm are collected in \autoref{tab:table2}, which are resolved after running a number of simulations. Meanwhile, the configuration of the neural networks is determined as follows: the filter number of layer "Conv\_1\_1" and "Conv\_1\_2" are 8 and 16, respectively. For each convolutional layer, the kernel size is set to 3, while for layer "Conv\_1\_2" an one-dimension padding operation is added to support the convolutional calculation. In addition to the convolutional layers, the kernel sizes of max-pooling layers are all set to 2 with a 2-step stride, while an one-dimension padding operation is also attached to the second max-pooling layer. The configuration of layer "Conv\_2\_1" and "Conv\_2\_2" is the same as those in action parameter network. The number of neurons for "Dense\_1\_1" and "Dense\_1\_2" is set to 128 and 64 , and for "Dense\_2\_1" and "Dense\_2\_2" is 256 and 64.

\begin{table}[htb]\caption{Hyperparameters of the PRL algorithm}
\centering
\label{tab:table2}
\footnotesize
\begin{threeparttable}
\begin{tabularx}{\textwidth}{p{0.4\textwidth}YY}
\toprule
Hyperparameter & Symbol & Value \\
\midrule
Discount factor & $\gamma$ & 0.99 \\
Averaging rate for $Q$ & $\tau_1$ & $10^{-2}$ \\
Averaging rate for $\mu$ & $\tau_2$ & $10^{-3}$ \\
Step size for $Q$ & $\alpha$ & $10^{-4}$ \\
Step size for $\mu$ & $\beta$ & $10^{-5}$ \\
Initial exploration factor & $\epsilon_{init}$ & 1 \\
Terminal exploration factor & $\epsilon_{end}$ & $10^{-2}$ \\
The number of decaying episodes & $n_{step}$ & $10^3$ \\
Minibatch size & $\mathcal{B}$ & 128 \\
Replay buffer size & $\mathcal{R}$ & $5\times10^5$ \\
\bottomrule
\end{tabularx}
\end{threeparttable}
\end{table}

\section{Simulation and discussion}
\label{sec:result}
\subsection{Simulation configuration}
The simulation environment is built in SUMO \citep{SUMO2018} to test the performance of both training phase and evaluation phase. As an highly portable microscopic traffic simulation platform, SUMO offers abundant pre-defined driver models, and meanwhile the operation of vehicles can be controlled through Traci interface by Python programs. Therefore, we delimit the HDVs and CVs by adding PRL policy programmed in Python module to CVs, while the operation of HDVs is based on the default Krauss car-following model and LC2013 lane-changing model. The energy model that produces transient electricity consumption is derived from \citep{kurczveil2013}, which leverages a braking energy recovery mechanism. It means that the value of instantaneous energy consumption $e_t^i$ is negative when the vehicle perform a braking operation, which presents challenges to searching a globally optimal eco-driving policy. The specific form and parameters of the energy model are given in our previous article \citep{TRR}. The proposed PRL policy do not need to explicitly analysis the energy model, while it only receive a transient signal that indicates the energy consumption in a model-agnostic manner.

\begin{figure}[htp]
    \centering
    \includegraphics[width=0.95\textwidth]{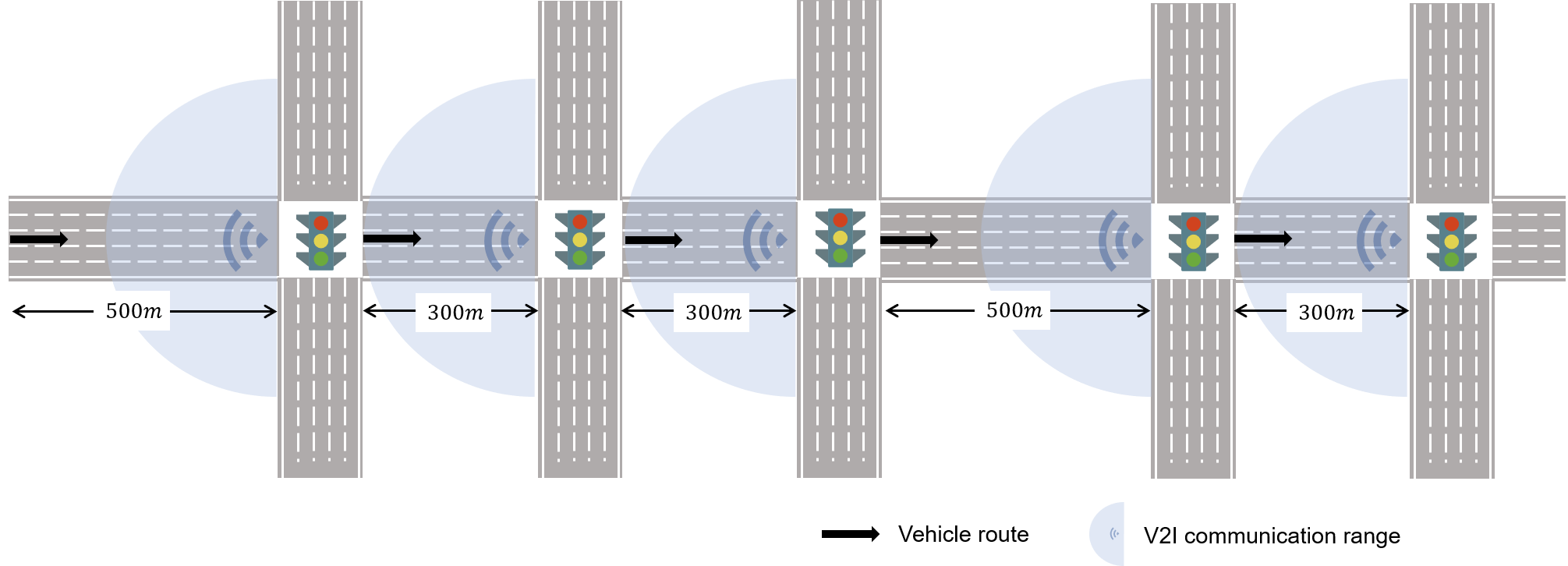}
    \caption{The road network scenario used for simulation.}
    \label{fig:env}
\end{figure}

We create a coordinated signal control scenario and a non-coordinated signal control scenario in a multi-intersection environment to evaluate the proposed framework. As illustrated in \autoref{fig:env}, the corridor contains five consecutive signalized intersections, for which a V2I communication range is set to $300m$. The traffic lights with cycle time of $90s$ change synchronously for non-coordinated scenario, while for coordinated scenario a series of offset is set up to ensure that a vehicle can encounter a green wave under free flow condition.

The parameters related to the simulation follow those in \autoref{tab:table3} if no otherwise specified. Besides, the values of the penalty items is set as $r_p^s=-30$, $r_p^j=-30$, and $r_p^c=-50$. Together with the weighting parameter $\varphi$, the reward-related parameters will be investigated in sensitivity analyses.

\begin{table}[htb]\caption{Parameter configuration for the simulations}
\centering
\label{tab:table3}
\footnotesize
\begin{threeparttable}
\begin{tabularx}{\textwidth}{p{0.4\textwidth}YYY}
\toprule
Simulation parameter & Symbol & Value & Unit \\
\midrule
Maximum deceleration & $a_{min}$ & -4 & $m/s^2$ \\
Maximum acceleration & $a_{max}$ & 3 & $m/s^2$ \\
Road speed limit & $V_{max}$ & 50 & $km/h$ \\
Cell length & $L_c$ & 1 & $m$ \\
Forward perceived range & $L_{head}$ & 50 & $m$ \\
Backward perceived range & $L_{tail}$ & 10 & $m$ \\
Vehicle length & - & 5 & $m$ \\
Lane changing duration & $T_c$ & 3 & $s$ \\
Velocity penalty threshold & $v_{\zeta}$ & 1.5 & $m/s$ \\
Jerk penalty threshold & $j_{\zeta}$ & 4 & $m/s^3$ \\
\bottomrule
\end{tabularx}
\end{threeparttable}
\end{table}

In addition to the general parameter settings of the simulation platform, we still need to take measures to provide a more dynamic traffic condition, with the intent to train the agent efficaciously. On one hand, traffic flow with $1000veh/h$ is generated in the simulation environment in a random departure mode, which aims at modelling a random road traffic situation; on the other hand, the ego vehicle is inserted to the beginning of the road network in a random fashion during training period. More precisely, the insertion time of the ego vehicle adheres to an uniform distribution $U(150s,300s)$, making the agent encounter different traffic signal status and vehicle queue situations, which is of benefit to enhance the robustness of the algorithm.

\subsection{Training phase discussion}
At first, we discuss about the settings of the reward function. The study investigate not only the forms of the reward function (i.e., $r_t^1$, $r_t^2$, and $r_t$), but also the impact of weighting parameter $\varphi$, to analyze the optimal reward configuration by empirical experiments. It should be noted that we also add penalty $r_p^c$ for $r_t^1$ and $r_t^2$ to prevent repeated lane changing behaviors. 

\begin{figure}[!h]
\begin{subfigure}{\textwidth}
  \centering
  \includegraphics[width=1\textwidth]{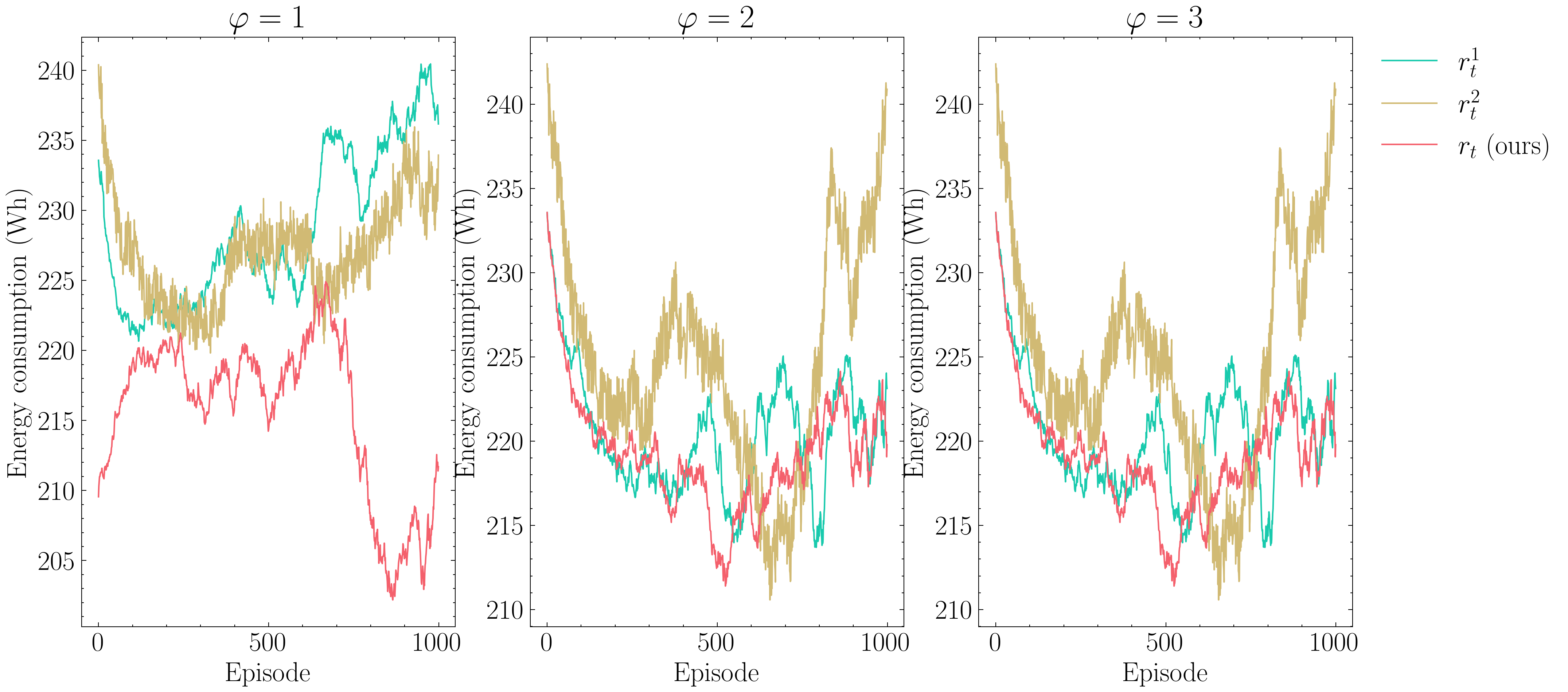}
  \caption{}
\end{subfigure}
\begin{subfigure}{\textwidth}
  \centering
  \includegraphics[width=1\textwidth]{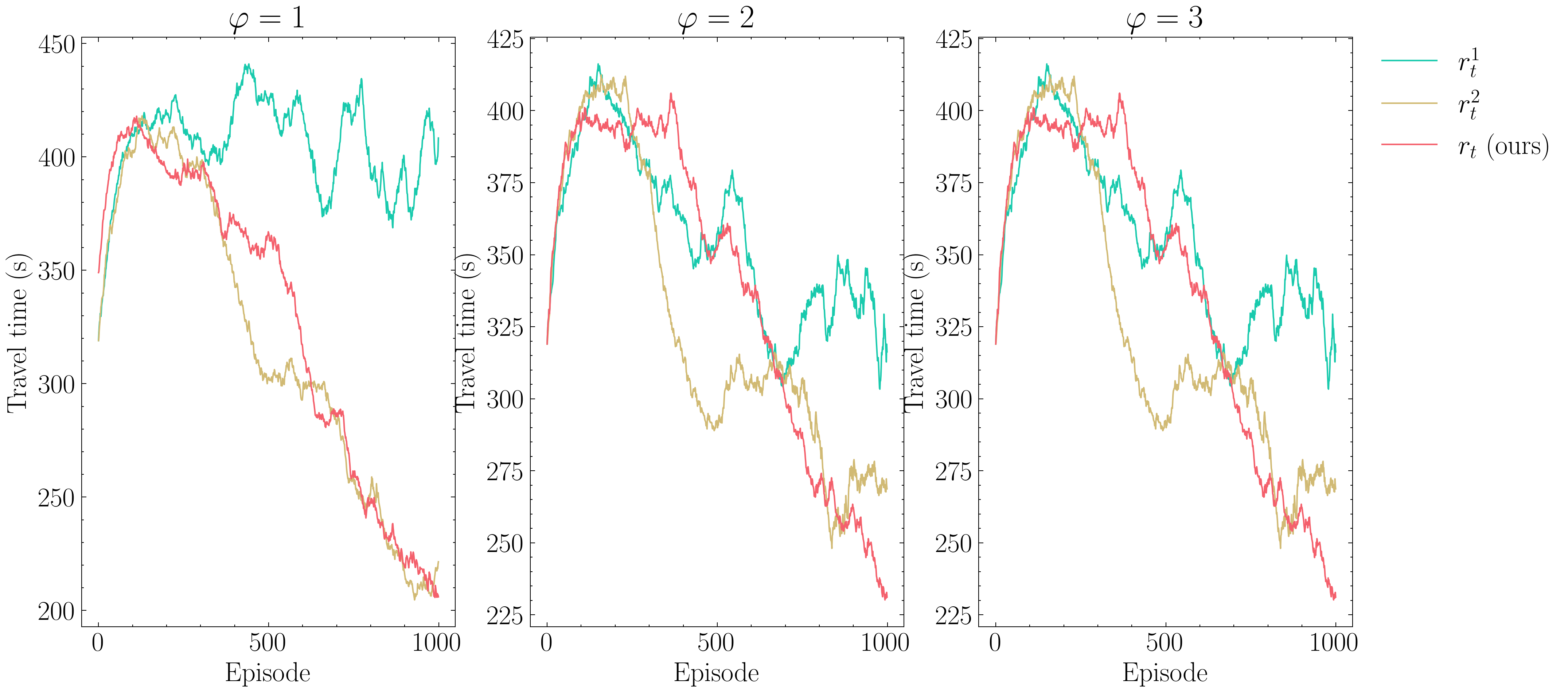}
  \caption{}
\end{subfigure}
\caption{The changing trend of corresponding indicators during training period. The data collected in episode $D_k$ is smoothed by $D_k \leftarrow 0.98D_{k-1} + 0.02D_k$. (a) Energy consumption, (b) Travel time.}
\label{fig:rwd_compare}
\end{figure}

\autoref{fig:rwd_compare} compares three reward configurations in terms of the energy consumption and travel time for the ego vehicle throughout the 1000-episode training period. Different patterns are seen in the variation of energy and time consumption for $r_t^1$ and the other reward settings. No matter how the parameter $\varphi$ changes, the downward trend of the metrics is not significant for $r_t^1$, illuminating that the agent cannot learn effective information with feedback only at the end of the episodes. In contrast, the converged energy consumption of $r_t^2$ and our $r_t$ are much lower, but the gaps between the those two rewards are likely to widen. In addition to the energy consumed by the ego vehicle, the changing trend of time consumption of both $r_t^2$ and $r_t$ are similar, but the figures of $r_t$ see a relatively stable pattern.

On the other side, the impact of the weighting parameter $\varphi$ is likely to be insignificant, because the results remain basically unchanged with the increase of $\varphi$. It is possibly due to that the optimal driving patterns are identical when we give similar importance to mobility and energy saving. As a result, we will use $\varphi=1$ in the rest of the simulations since the numerical magnitudes of the two indicators are close.

\begin{figure}[htb]
\begin{subfigure}{0.5\textwidth}
\centering
  \includegraphics[width=\textwidth]{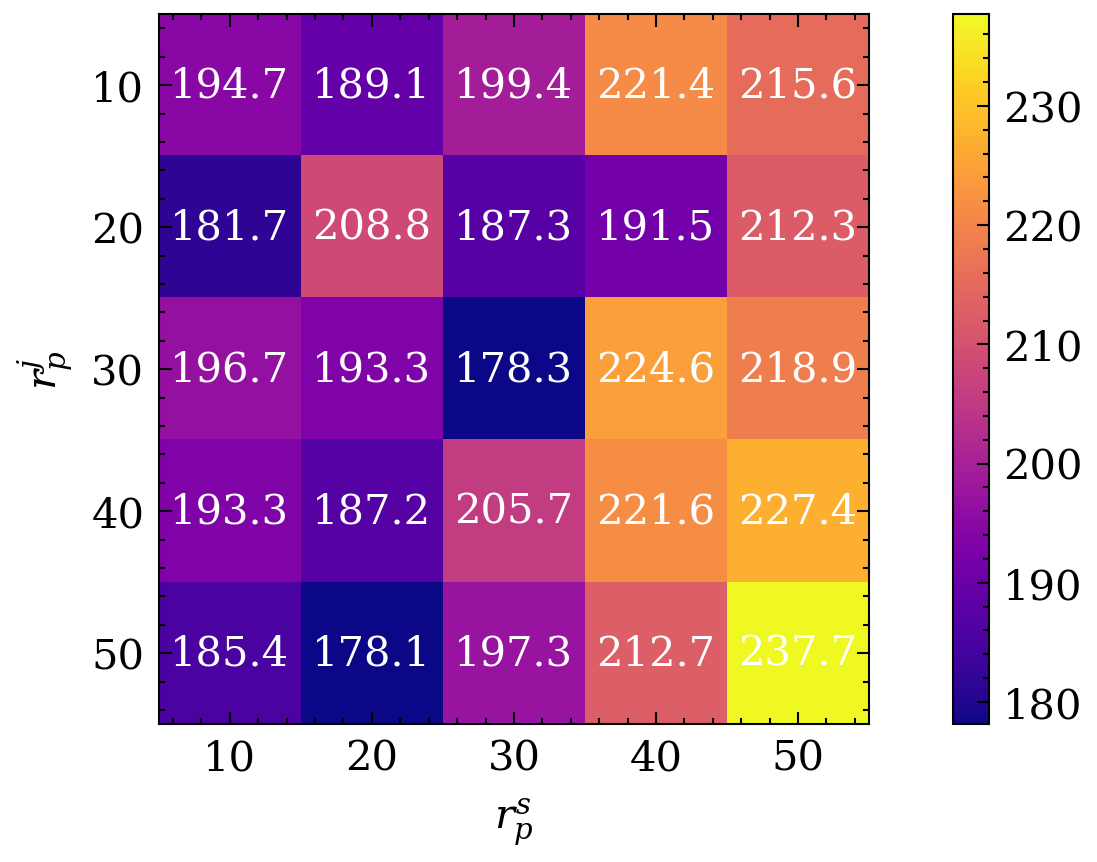}
  \caption{}
\end{subfigure}
\begin{subfigure}{0.5\textwidth}
\centering
  \includegraphics[width=\textwidth]{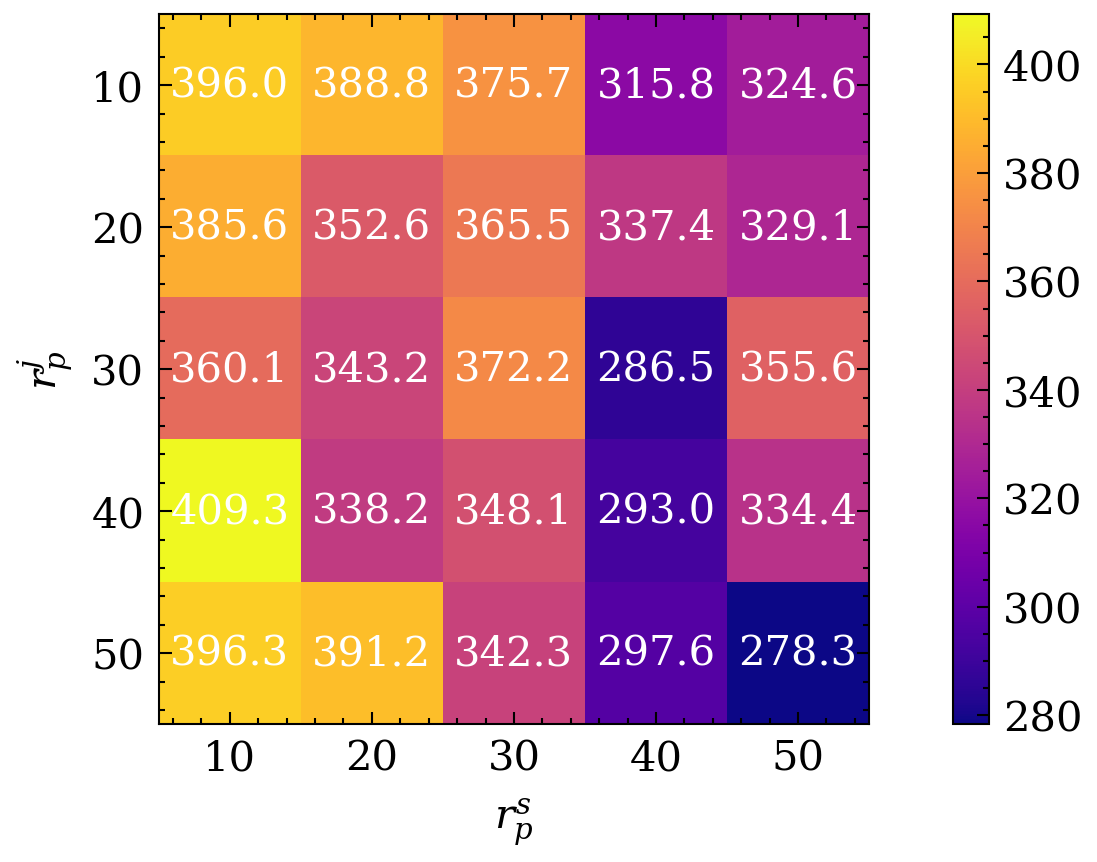}
  \caption{}
\end{subfigure}
\caption{The heatmap with different value of $r_p^s$ and $r_p^j$. (a) Energy consumption, (b) Travel time.}
\label{fig:heatmap}
\end{figure}

With regard to the values of $r_p^s$, $r_p^j$, and $r_p^c$, a series of simulation-based training are carried out to explore the optimal configuration. We fix $r_p^c=50$ and vary $r_p^s$ and $r_p^j$ from 10 to 50 with $step=10$ on the grounds that the result may be changed by different relative ratio of the three items. The simulation investigation is based on the environment with non-coordinated traffic signal setting, since this scenario can be more challenging. The results are shown in \autoref{fig:heatmap}, which utilizes average value among 100 test cases for each setting. According to the graph, we find that the growth of $r_p^S$ can significantly reduce travel time but increase the energy consumption. This is reasonable because the vehicle may speed up as soon as possible if no sufficient penalty is imposed on its reward system, which also resembles some human drivers. By comparison, the variation of $r_p^j$ seems to have much lower impact on travel time and electricity consumption, but it helps to smooth the trajectory of the CV, as negative reward will be imposed if the vehicle perform violent acceleration or deceleration. As a result, we choose $r_p^s=40$ and $r_p^j=30$ to keep a trade-off between energy efficiency and traffic mobility.

\subsection{Overall performance of single vehicle}

After obtaining the well-trained models of the agent, we can deploy them in new environment to test the performance of the proposed method. We firstly introduce a baseline that embedded in SUMO to verify if our model can outperform human-driving behaviors. In this case, there is no difference between CVs and HDVs, whereas the CVs follow the default Krauss car-following model and the LC2013 lane-changing model to simulate the driving behaviors of human drivers. Meanwhile, we deploy the well-trained agent to the environments with the same insertion time of the ego vehicle, and record the data of trajectory, speed, and cumulative energy consumption. 

\begin{figure}[!h]
  \includegraphics[width=1\textwidth]{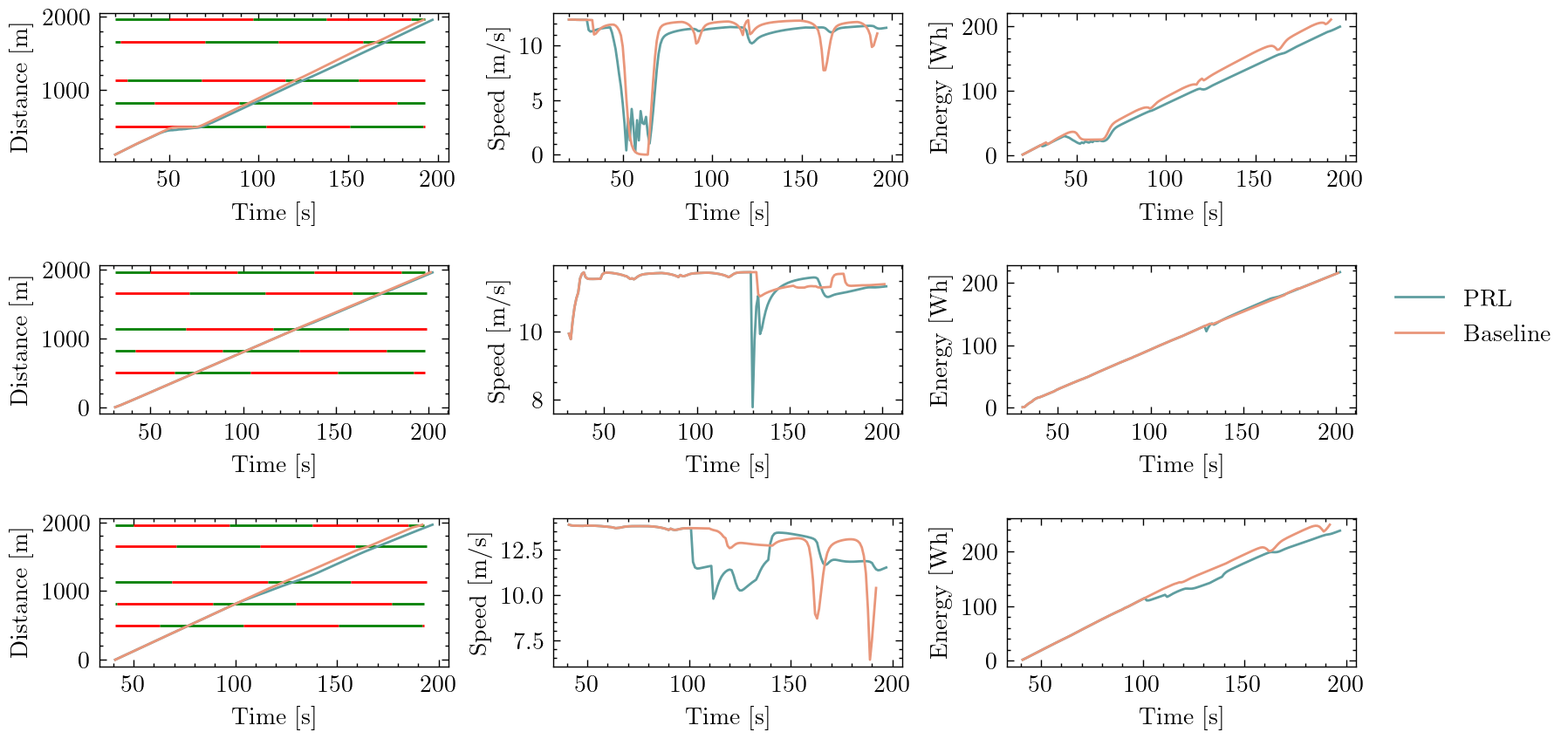}
\caption{The comparison of baseline and PRL in coordinate signal environment.}
\label{fig:compare1}
\end{figure}

\begin{figure}[!h]
  \includegraphics[width=1\textwidth]{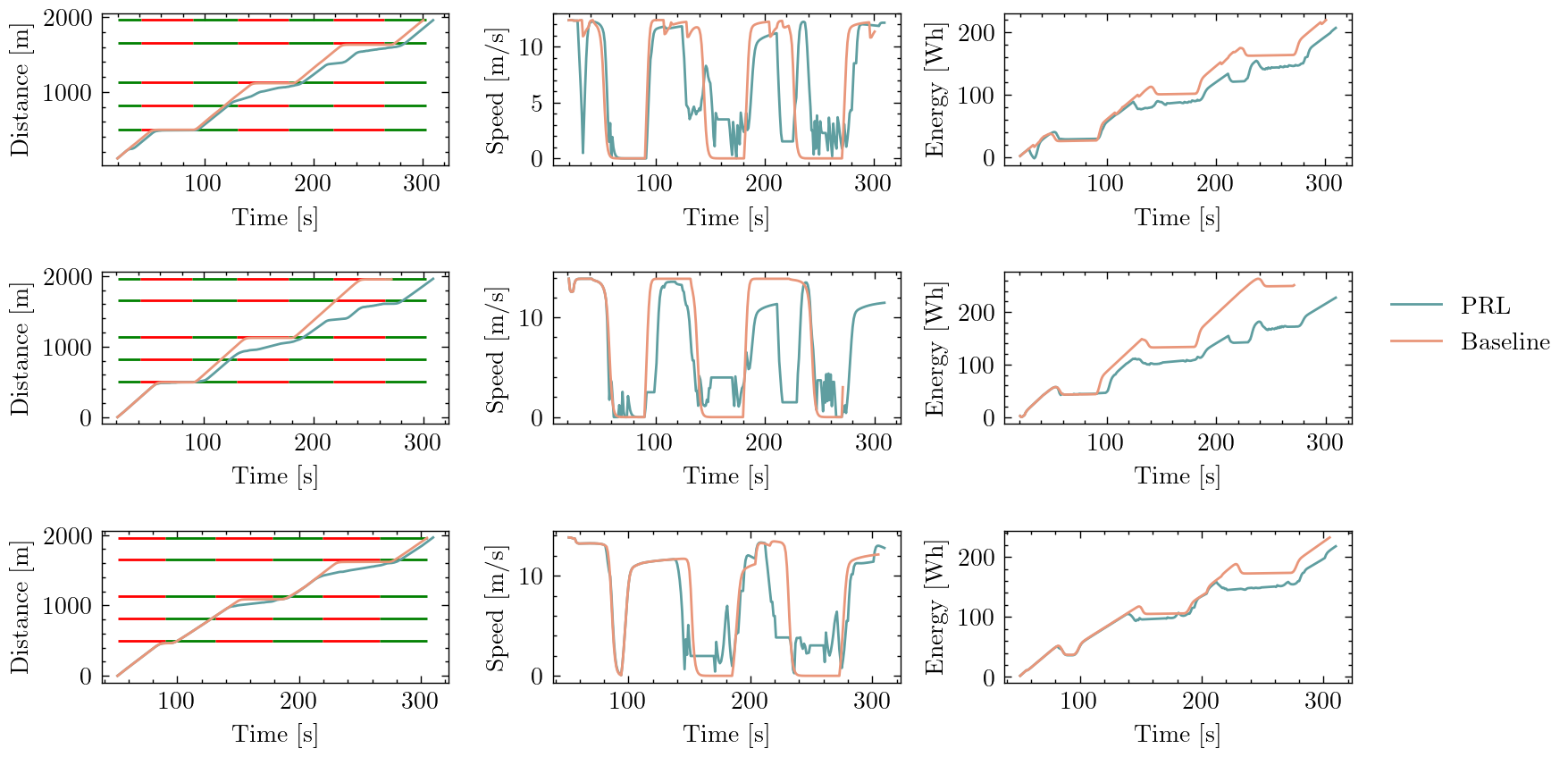}
\caption{The comparison of baseline and PRL in non-coordinate signal environment.}
\label{fig:compare2}
\end{figure}

For both coordinate and non-coordinate signal environment, we generate three scenarios respectively to test the method and make visually comparisons, and the results are illustrated in \autoref{fig:compare1} and \autoref{fig:compare2}. It can be found that the agent trained by PRL can adapt to different traffic environment: (1) in coordinate traffic signal environment, the speed profiles of PRL-based vehicle are similar to those based on human-driving model, but it learns a strategy of early deceleration to decrease energy consumption. (2) in non-coordinate traffic signal environment, the energy-efficiency of the vehicle can be better reflected. While the vehicle with baseline model is obliged to stop at every red signal, the CV controlled by PRL policy can improve its energy-efficiency by early deceleration and braking-reacceleration in small range to recover the electricity through the braking-recovery mechanism. In addition to the energy indicator, the increase of travel time is not significant for PRL policy. Due to the fact that the vehicle with default model always drives in its maximum acceptable velocity, it is difficult to improve mobility and energy-efficiency at the same time. Therefore, our learning-based framework is capable of reducing the consumed energy with similar traffic delay of baseline, which is much more practical in real situations.

In addition, 100 test cases are simulated to compare the general performance of the proposed algorithm. \autoref{tab:single} presents the overall performance of the PRL algorithm. Compared with the manual driving behavior, although a fraction of travel time is sacrificed, the proposed framework can significantly reduce the electricity consumption, showing a decrease of 4.1\% in coordinate signal environment and 27.13\% in non-coordinate signal environment.

\begin{table}[htb]\caption{Average performance of the PRL algorithm among 100 test cases}
\centering
\begin{tabular}{c|cccc}
\Xhline{1pt}
Method & \multicolumn{2}{c}{Energy consumption (Wh)} & \multicolumn{2}{c}{Travel time (s)} \\
\Xcline{2-3}{0.4pt} \Xcline{4-5}{0.4pt}
 & coord. & non-coord. & coord. & non-coord. \\
\Xhline{1pt}
Baseline & 231.13 & 253.79 & 189.08 & 260.21 \\
PRL & 221.75 & 184.93 & 204.79 & 289.82 \\
Imp. $\uparrow$ & 4.1\% & 27.13\% & -7.93\% & -11.25\% \\
\Xhline{1pt}
\end{tabular}
\label{tab:single}
\end{table}

In order to show the superiority of our method in a more general way, we make further comparison studies with the following strategies:
\begin{itemize}
    \item DDPG+LC2013: the longitudinal motions of the CVs are expected to be controlled by a DDPG policy with the same MDP configurations of our PRL (the harmony coefficient $r_p^c$ is excluded), while the lateral decisions are based on the classical LC2013 model.
    \item DQN: the CVs are controlled by a DQN policy in this case. Due to that the value-based DRL algorithm can only handle with discrete action space, a predefined longitudinal action set is given as:
    \begin{equation*}
    \begin{split}
            A^{DQN}_{lon}=\{1.0a_{max}, 0.75a_{max}, 0.5a_{max}, 0.25a_{max}, 0.0, \\ 0.25a_{min}, 0.5a_{min}, 0.75a_{min}, 1.0a_{min}\}
    \end{split}
    \end{equation*}
    Combining with the lateral action set $A^{DQN}_{lat}=\{-1,0,1\}$ by Cartesian product, we get a 27-dimensional vector to represent the action space.
\end{itemize}

With the intent to make a fair comparison, the hyperparameters for DDPG and DQN, such as $\gamma$, $\alpha$, $\mathcal{B}$, and $\mathcal{R}$ are the same as our PRL, as well as the general architectures of the neural networks. The exploration mechanism of DDPG and DQN are identical to the OU process noise and decaying $\epsilon$-greedy strategy in the PRL algorithm, respectively. 

\begin{figure}[!h]
\begin{subfigure}{0.5\textwidth}
\centering
  \includegraphics[width=\textwidth]{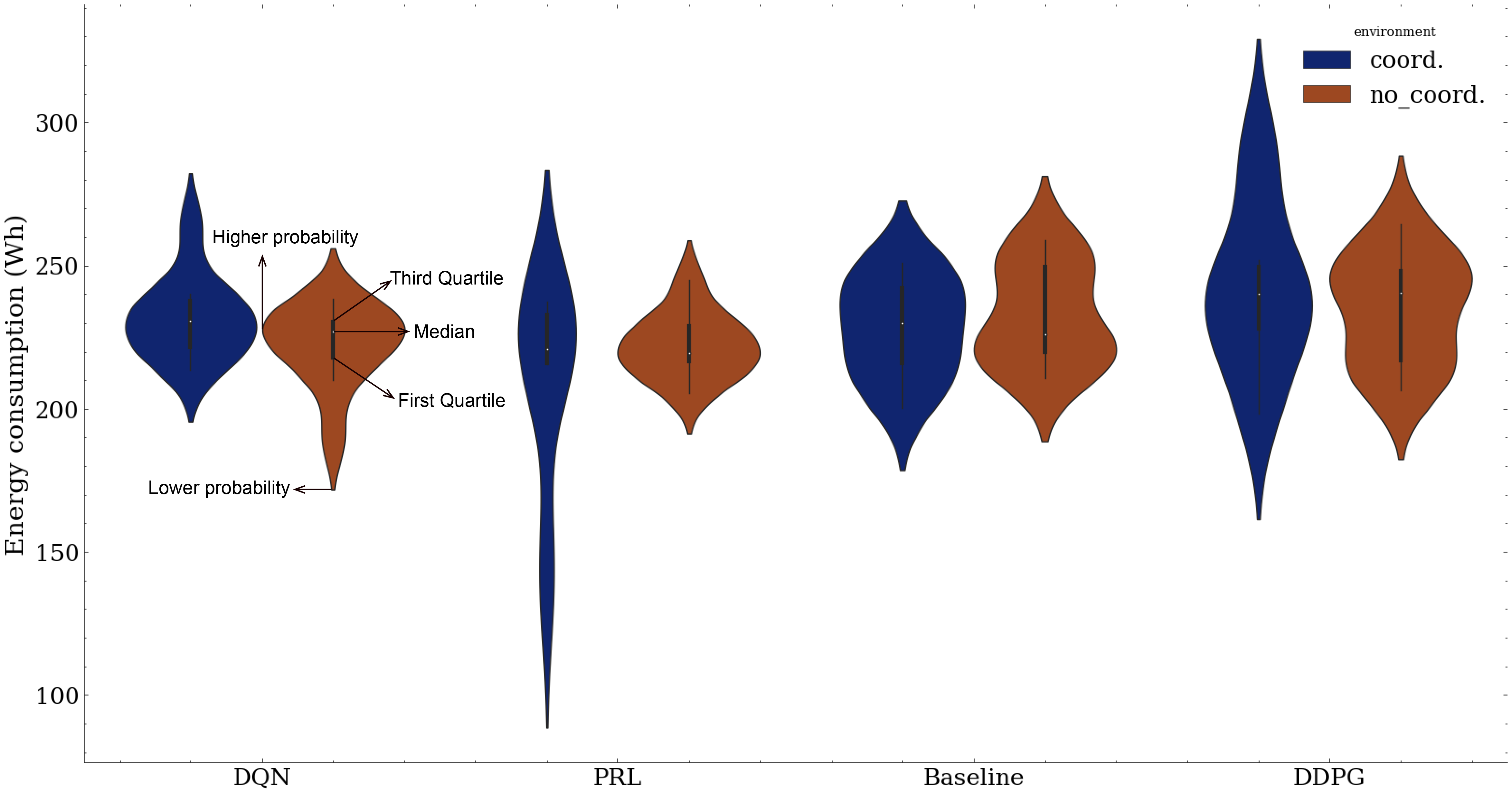}
  \caption{}
\end{subfigure}
\begin{subfigure}{0.5\textwidth}
\centering
  \includegraphics[width=\textwidth]{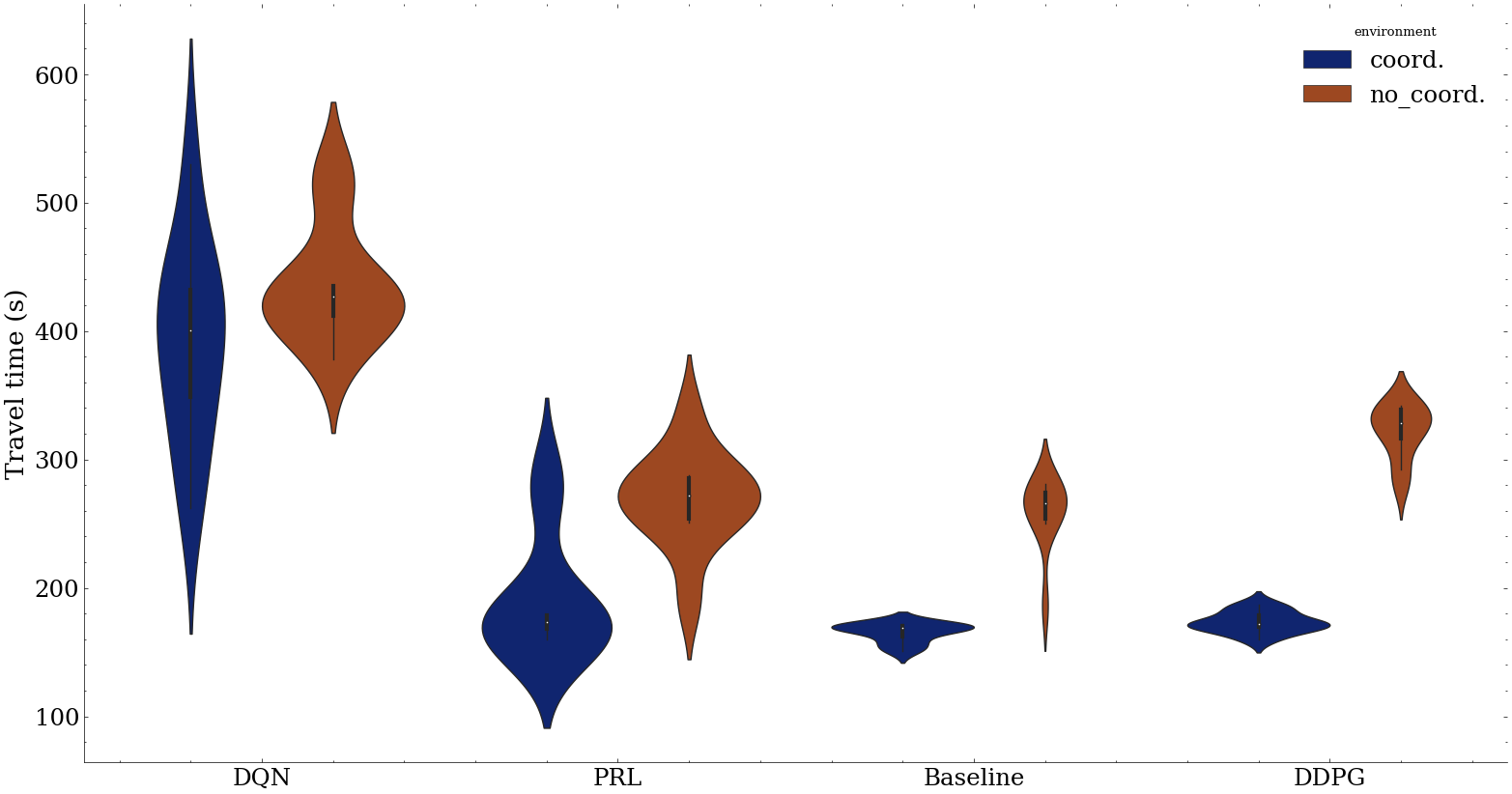}
  \caption{}
\end{subfigure}
\caption{The performance of the ego vehicle with different control methods. (a) Energy consumption, (b) Travel time.}
\label{fig:rwd_bar}
\end{figure}

To alleviate the impact caused by randomness, the results are collected in 10 independent group of simulations with different value of random seed. \autoref{fig:rwd_bar} shows the collected data in a violin plot form, which depicts summary statistics and the density of the results. From the figure, we can draw following conclusions: (1) DQN cannot satisfactorily be applied in the task, whereas it sees an unacceptable performance in terms of travel time. This can be caused by the large discrete action space of the DQN agent when we incorporating longitudinal and lateral decisions. (2) The performance of DDPG is close to the baseline approach, but much worse in non-coordinate signal settings due to the increased travel time. (3) In contrast, the proposed PRL method is more impressive with dynamic traffic flow and signal changing status. Compared with other DRL-based methods, the distribution of energy consumption for PRL agent is close to a smaller value, with only a modest growth of travel time.

\subsection{Lateral performance of single vehicle}
\begin{figure}[!h]
\centering
  \includegraphics[width=0.8\textwidth]{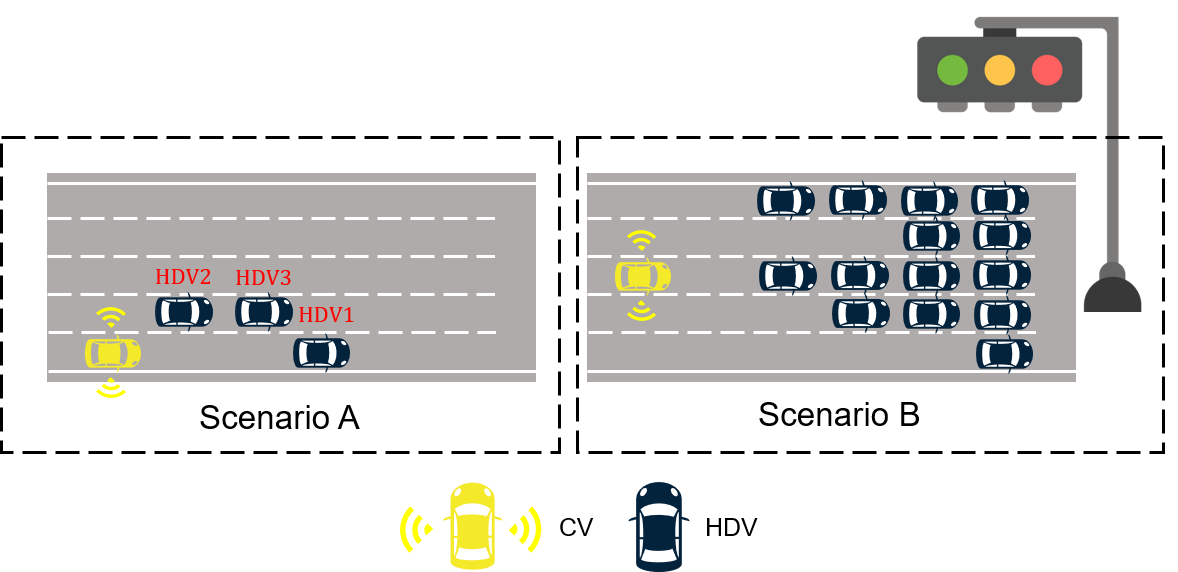}
\caption{The scenarios to test the lateral decision of the agent.}
\label{fig:scenario}
\end{figure}

\begin{figure}[!h]
\centering
\begin{subfigure}{0.47\textwidth}
\centering
  \includegraphics[width=\textwidth]{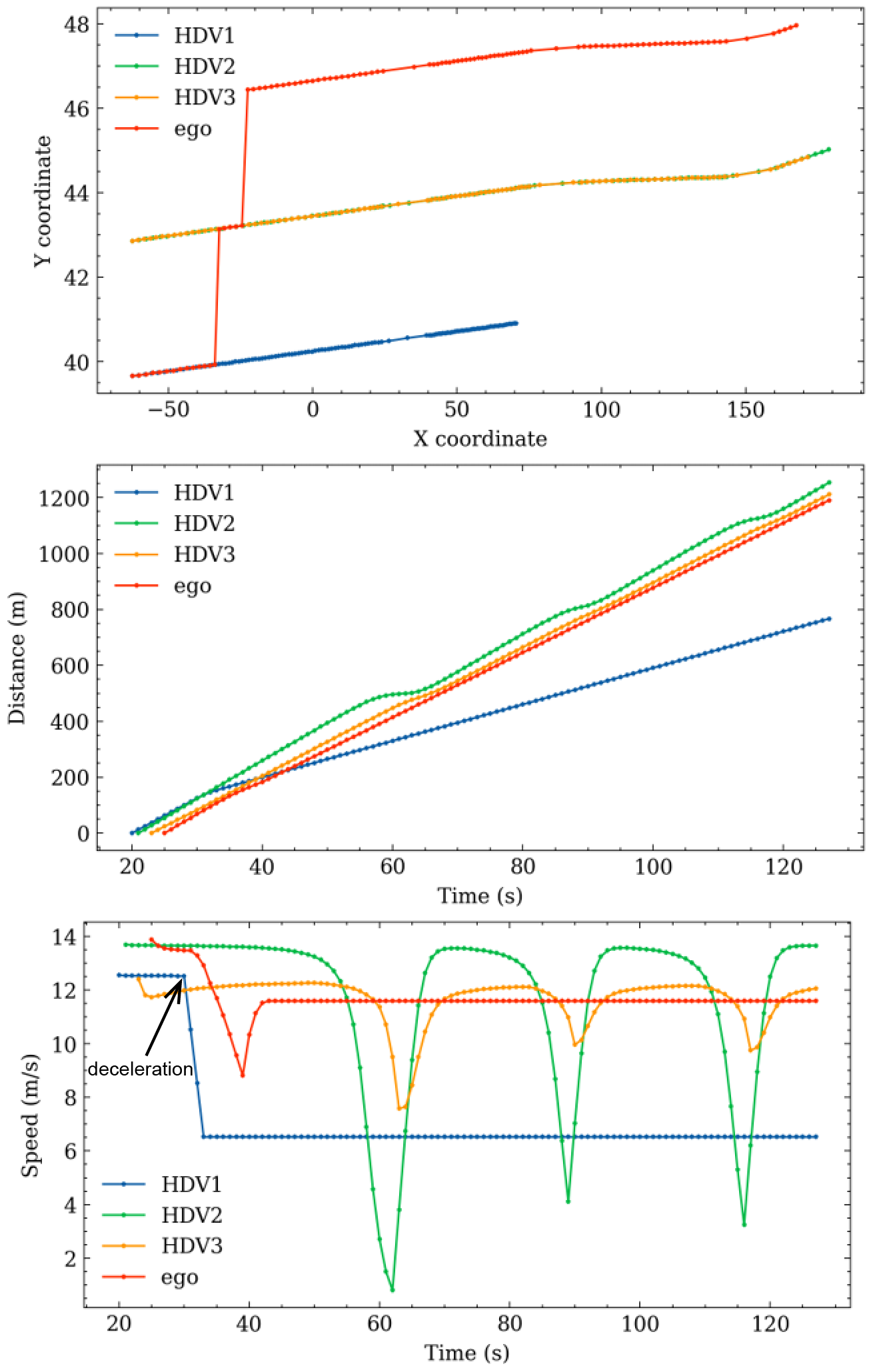}
  \caption{Scenario A}
\end{subfigure}
\begin{subfigure}{0.47\textwidth}
\centering
  \includegraphics[width=\textwidth]{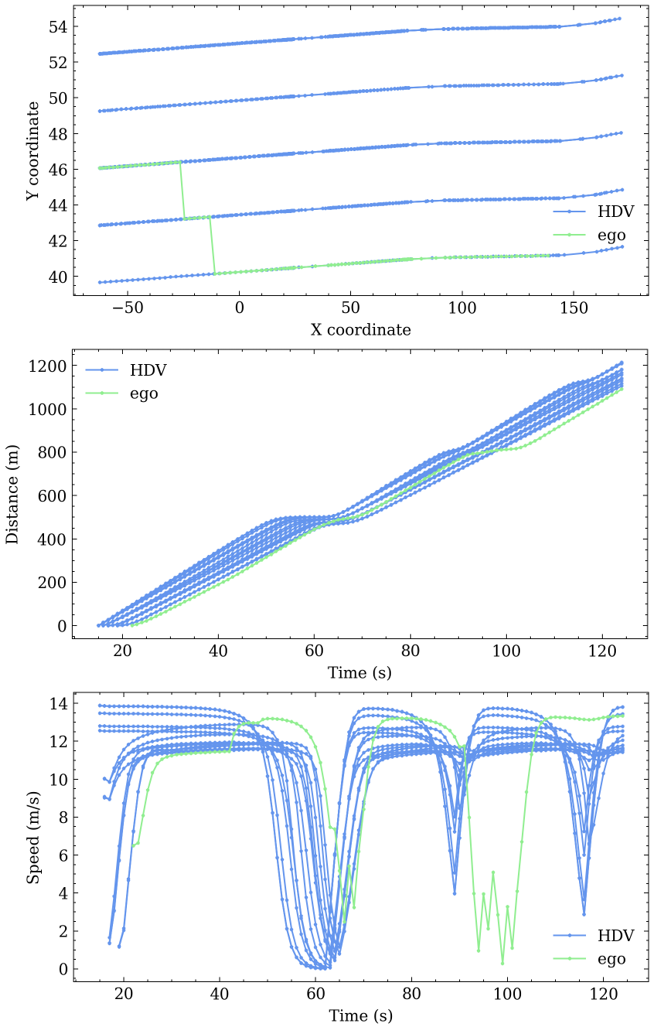}
  \caption{Scenario B}
\end{subfigure}
\caption{The performance of the ego vehicle in two synthetic scenarios.}
\label{fig:scenario_compare}
\end{figure}

It is difficult to illuminate the benefit of lane changing accurately as this benefit is not instantaneous. One feasible way to demonstrate the effectiveness of proper lateral decisions is to observe the behavior of the agent on some specific occasions. Thus, we take two synthetic scenarios as examples from a microscopic perspective and demonstrate that the agent is capable of making adaptive lateral decisions. As shown in \autoref{fig:scenario}, scenario A presents a situation that ego vehicle follows HDV1, which will be imposed a braking with deceleration of $2m/s^2$ in three consecutive time steps; scenario B depicts the ego vehicle follows a platoon composed of HDVs. As the queues of HDVs at the intersection is detected by the agent, it is supposed to change lane and plan its velocity in order to mitigate the interruption of the queue.

\autoref{fig:scenario_compare} depicts the simulation results in coordinate graphs, the time-distance graphs, and the time-speed graphs of scenario A and scenario B. \autoref{fig:scenario_compare}(a) shows the lane changing behaviors of the ego vehicle after HDV1 performing deceleration. Due to the existence of HDV2, the agent choose to decelerate and then change to the second lane. In order to shun the interruption of HDV2, it immediately change to the third lane and keep a smooth speed profile to cross the intersections. \autoref{fig:scenario_compare}(b) shows that the agent can select the lane with the least number of vehicles and slow down early to shun the potential influence of the leading HDV.

\subsection{Performance of mixed traffic}

\begin{figure}[!h]
\begin{subfigure}{0.5\textwidth}
\centering
  \includegraphics[width=\textwidth]{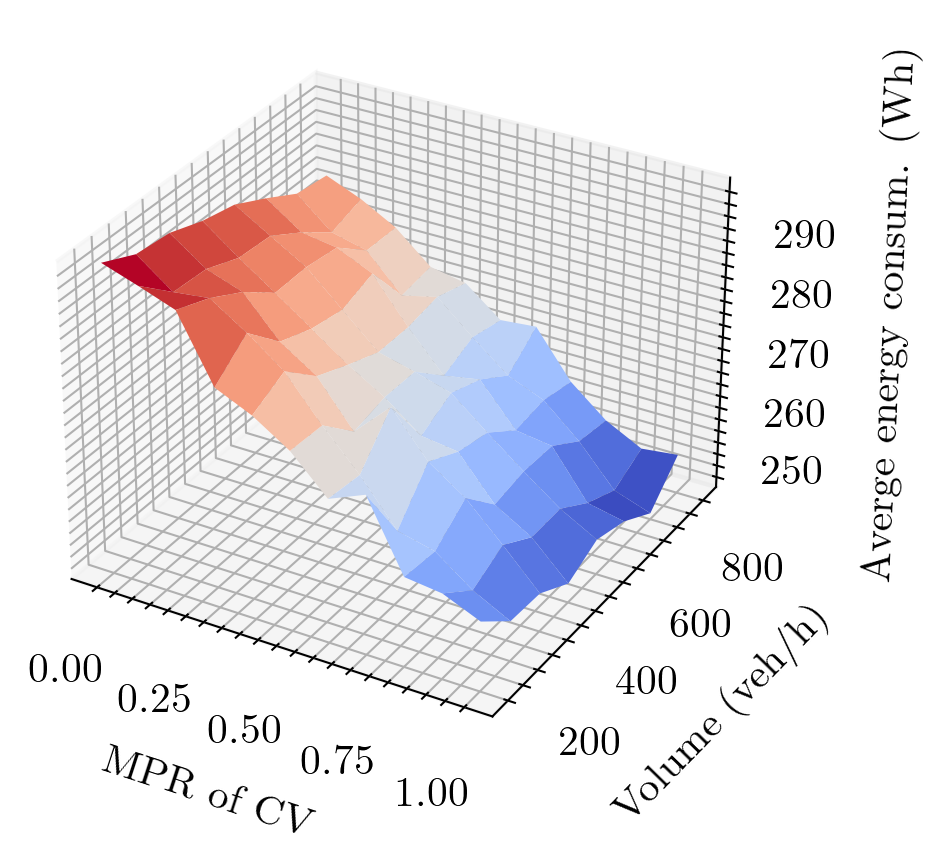}
  \caption{Average energy consumption (coordinate signal control)}
\end{subfigure}
\begin{subfigure}{0.5\textwidth}
\centering
  \includegraphics[width=\textwidth]{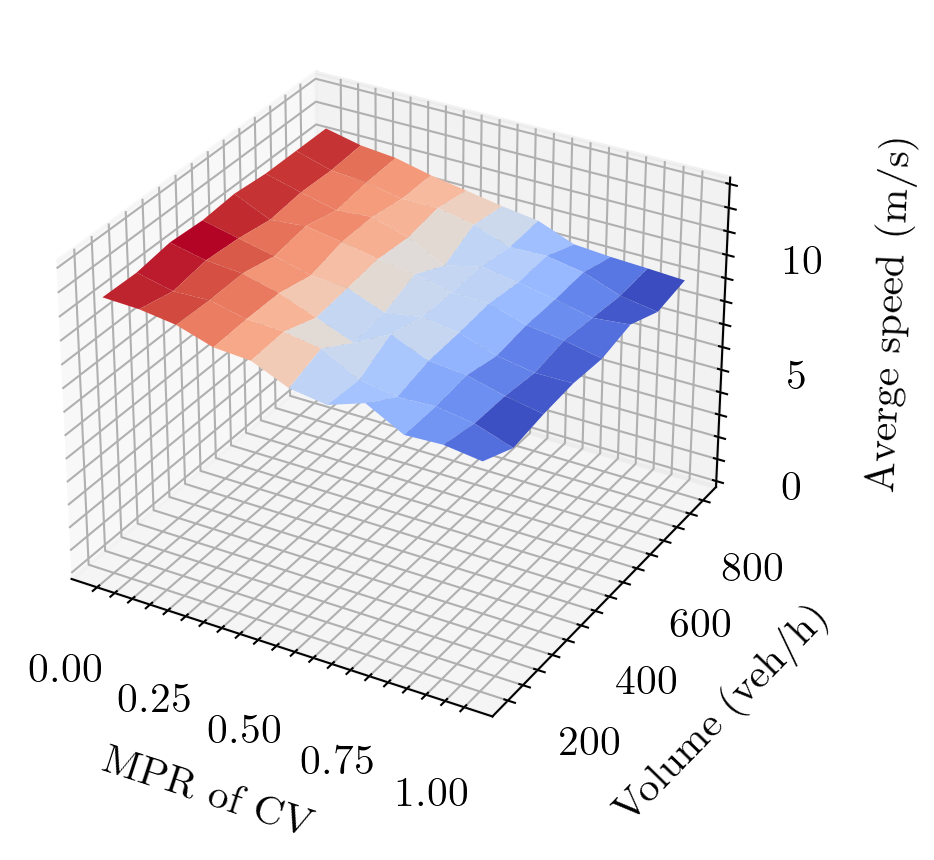}
  \caption{Average speed (coordinate signal control)}
\end{subfigure}
\begin{subfigure}{0.5\textwidth}
\centering
  \includegraphics[width=\textwidth]{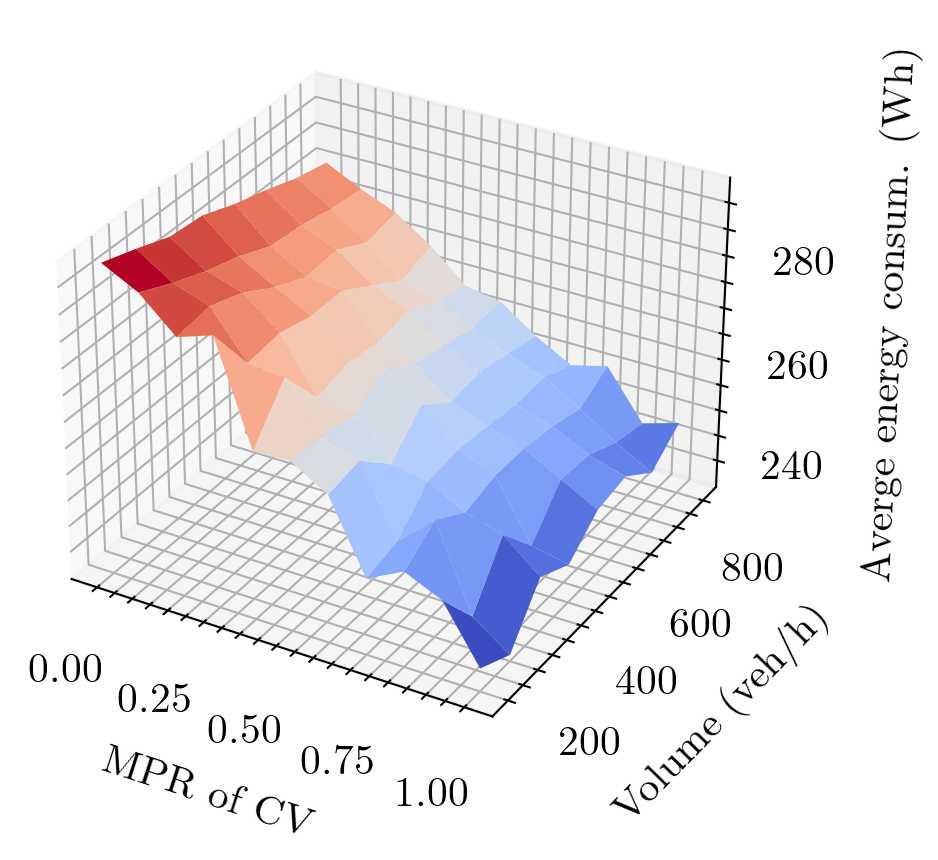}
  \caption{Average energy consumption (non-coordinate signal control)}
\end{subfigure}
\begin{subfigure}{0.5\textwidth}
\centering
  \includegraphics[width=\textwidth]{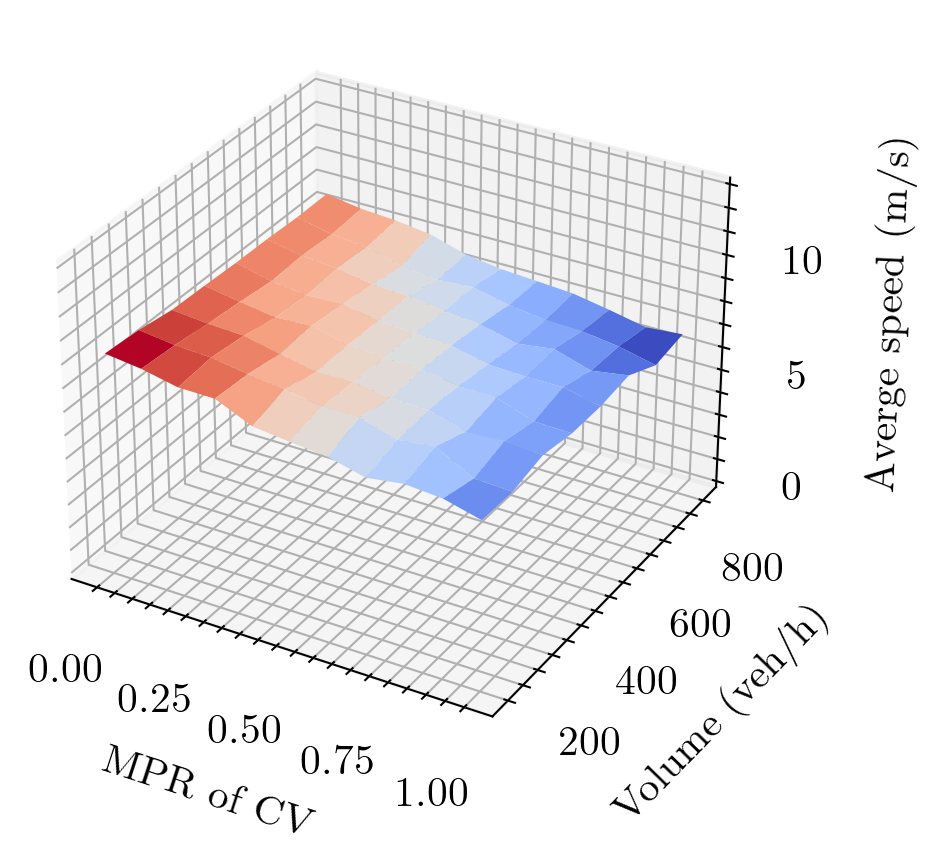}
  \caption{Average speed (non-coordinate signal control)}
\end{subfigure}
\caption{The performance of traffic with different traffic volume and CV MPR in different traffic signal environment.}
\label{fig:dd}
\end{figure}

Deploying the proposed strategy in mixed traffic flow is an essential part to verify the applicability of the study. Based on the belief that the co-existence of CVs and HDVs will last for many years, it is necessary to explore the potential impact of the strategy with different MPR of the CV.

\autoref{fig:dd} illustrates the traffic performance in terms of average electricity consumption and average speed with different MPR of CV and traffic volume. The time duration of single episode of simulation is set to $1800s$, and a group of simulation contains 5 independent experiments to collect data and calculate average results. It is obvious that marginal increase of MPR of CV will bring about relatively considerable energy reduction of the traffic. Compared with pure HDV scenario, the mean electricity consumption of scenario with CV MPR=1 decrease to 240Wh/veh-250Wh/veh, showing a decline of around 13.8\%. By deploying the PRL algorithm to the agent, the average energy consumption and average speed can maintain a stable value with different level of traffic demand when the MPR is fixed. Meanwhile, we can observe that the decrease of mean speed is not significant with the growth of MPR, demonstrating the sacrifice of vehicle mobility is negligible. Another finding is that the loss of mean speed in non-coordinate setting is lower than that in coordinate setting with the increase of MPR. This is due to the HDVs in non-coordinate signal environment are more likely to encounter red light and wait at the intersection. As such, the proposed method manages to transform this part of time to the duration that vehicles running with relatively lower speed to avert red traffic signals.

\begin{figure}[htb]
\begin{subfigure}{0.32\textwidth}
  \includegraphics[width=\textwidth]{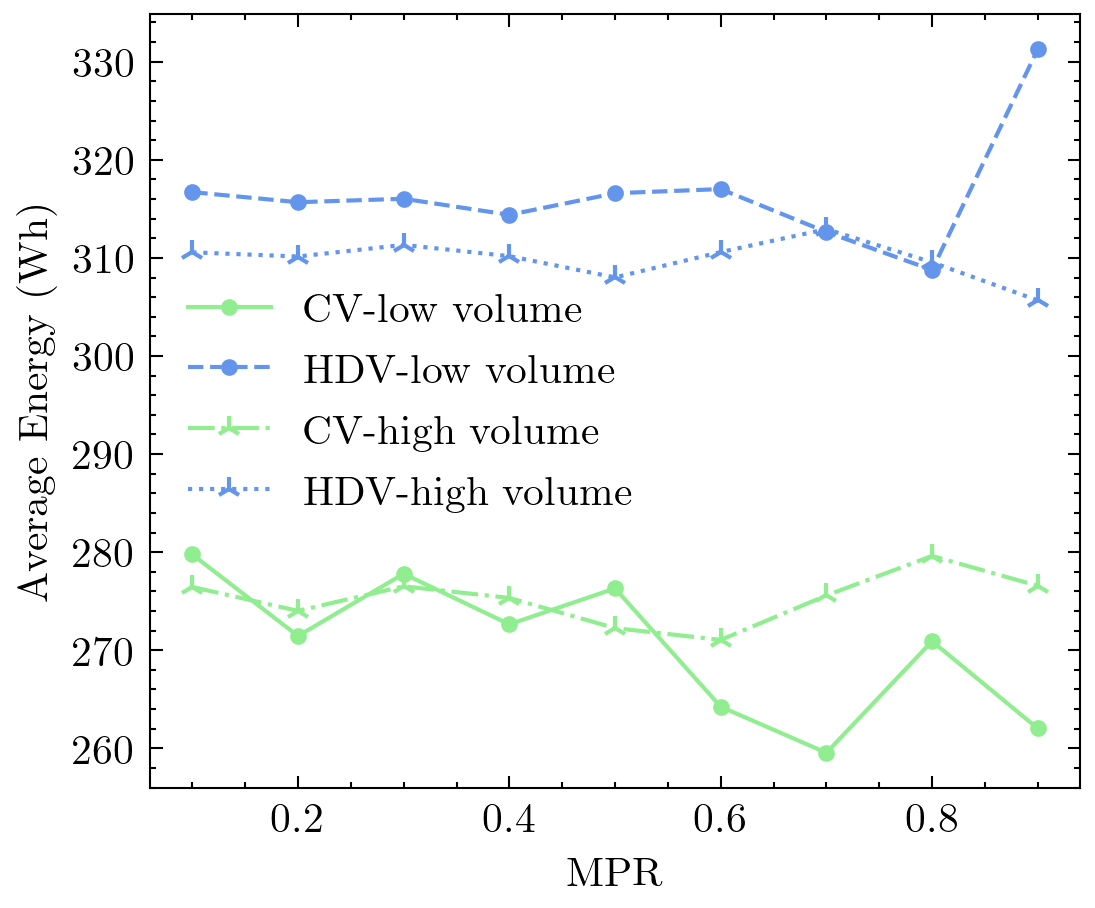}
  \caption{}
\end{subfigure}
\begin{subfigure}{0.32\textwidth}
  \includegraphics[width=\textwidth]{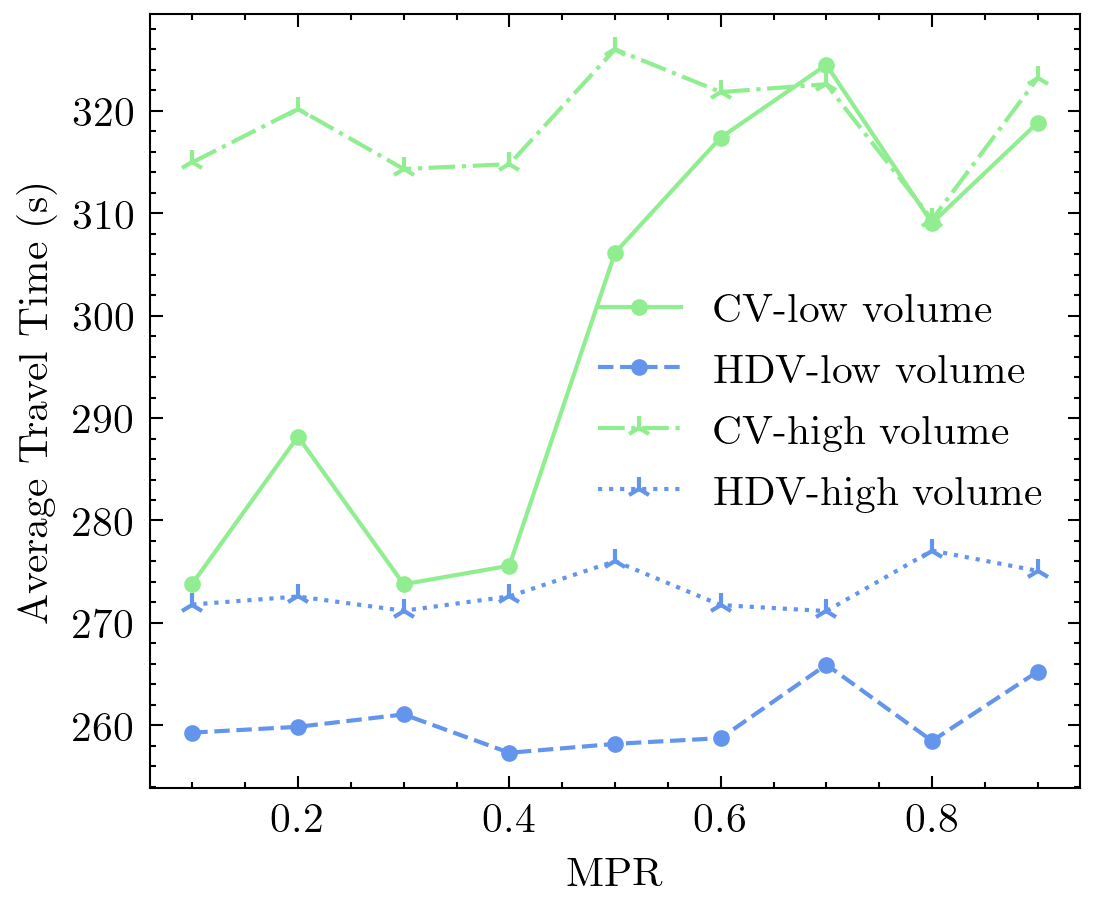}
  \caption{}
\end{subfigure}
\begin{subfigure}{0.32\textwidth}
  \includegraphics[width=\textwidth]{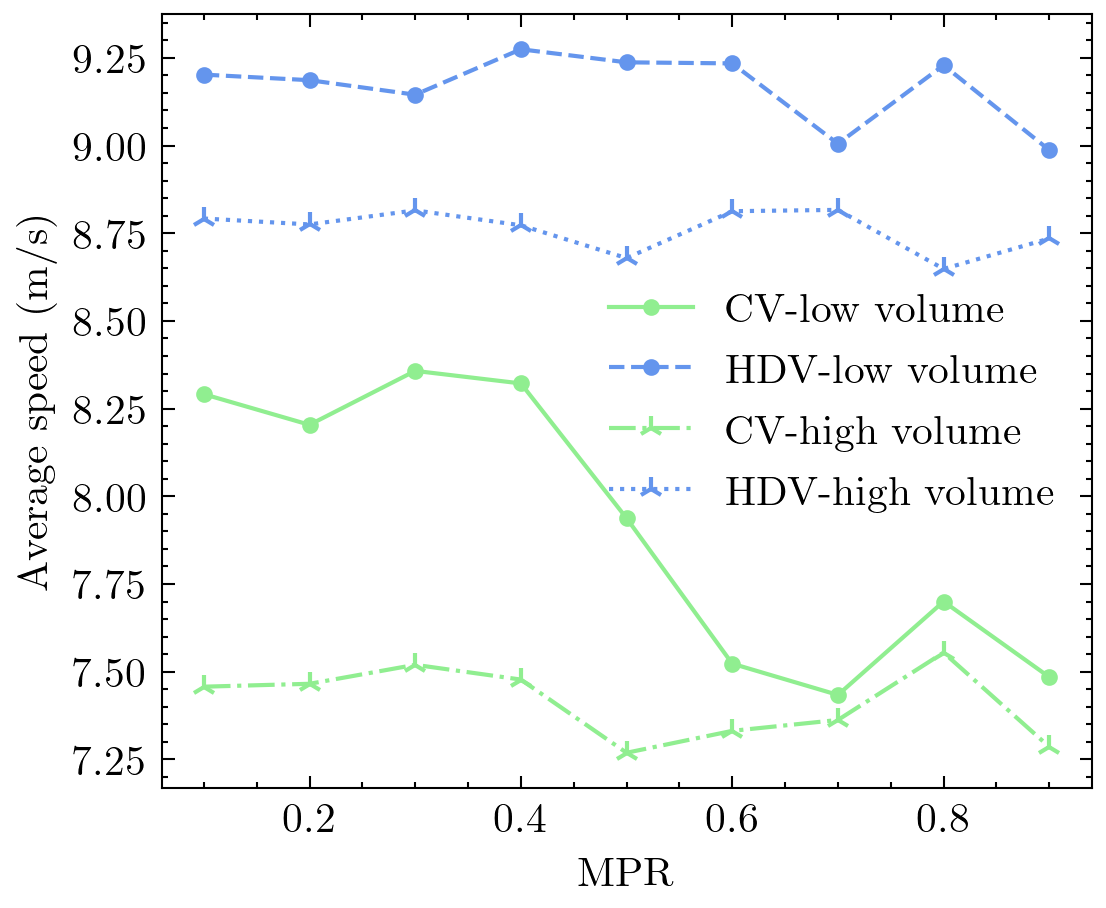}
  \caption{}
\end{subfigure}
\caption{Average performance of CVs and HDVs in different MPR and traffic volume. (a) Average energy consumption, (b) Average travel time, (c) Average speed.}
\label{fig:avg}
\end{figure}

Since many eco-driving strategies may impede the normal operation of HDVs so as to exert an adverse impact on traffic flow, we run simulations to observe that if the CVs with PRL strategy interfere with other vehicles. \autoref{fig:avg} categorise the vehicles into CVs and HDVs, and compare the metrics in non-coordinate signal settings. The situations of low traffic volume (200veh/h/lane) and high traffic volume (700veh/h/lane) are compared. The figures demonstrate that the operations of HDVs are not influenced by CVs, while thier metrics keep at a steady level, no matter how the volume and MPR of CVs changes.


\section{Conclusions}
\label{sec:conclusions}
Eco-driving is one of the potent approaches to improve the energy efficiency of the electric vehicles. This paper propose a PRL-based framework, which can cope with the vehicle control task with hybrid action space, to implement eco-driving in urban connected intersections. Incorporating with model-based car-following and lane-changing mechanism, the PRL algorithm trains the CV to learn to make joint decisions in terms of longitudinal dimension and lateral dimension. By analysing different reward settings, which are the core part of the DRL algorithm, this paper combines step-wise reward and terminal reward, and formulates an MDP model in a hybrid action space. 

Simulations are carried out in both coordinate signal setting and non-coordinate signal setting, with microscopic perspective and macroscopic perspective. The microscopic simulations investigate the performance of single agent in the dynamic traffic environment and show that the consumed energy can be reduced by 4.1\% with coordinate signal setting and 27.13\% with non-coordinate signal setting. Meanwhile, the PRL-based framework can outperform DQN and DDPG, which are used in some previous works to control CVs. The macroscopic simulations study the performance of the methodology when deploying the algorithm to multi-agent situations. It is found that the increase of CV MPR is of benefit to the overall performance of the traffic flow, while the operations of HDVs are not impaired by CVs with the PRL controller. These findings make the framework more practical when considering to apply DRL-aided policies in real-world intelligent vehicles.

Further exploration of the learning-based eco-driving strategies is very much needed, including the ways to consider the social welfare of the application of connected technologies and learning algorithms, so as to improve the performance of HDVs. In addition, the interactions between CVs is another area that deserves more attentions, which can promote cooperative eco-driving of CVs. By employing cooperative multi-agent DRL methods with specific communication protocols, the collaborative perception and cooperative decisions can be achieved to further dig the potential value of advanced eco-driving systems, especially under low CV MPR circumstances.

\section*{Disclosure statement}

The authors declare that they have no known competing financial interests or personal relationships that could have appeared to influence the work reported in this paper.

\section*{Funding}

This research was supported by the National Key R\&D Program of China (Grant 2021TFB1600500), and the Key R\&D Program of Jiangsu Province in China (Grant No. BE2020013).

\bibliographystyle{tfcad}
\bibliography{interactcadsample}

\end{document}